\documentclass[letterpaper]{article} % DO NOT CHANGE THIS
\usepackage{aaai2026}  % DO NOT CHANGE THIS

\usepackage{times}  % DO NOT CHANGE THIS
\usepackage{helvet}  % DO NOT CHANGE THIS
\usepackage{courier}  % DO NOT CHANGE THIS
\usepackage[hyphens]{url}  % DO NOT CHANGE THIS
\usepackage{graphicx} % DO NOT CHANGE THIS
\usepackage{natbib}  % DO NOT CHANGE THIS AND DO NOT ADD ANY OPTIONS TO IT
\usepackage{caption} % DO NOT CHANGE THIS AND DO NOT ADD ANY OPTIONS TO IT
\usepackage{algorithm}
\usepackage{algorithmic}

\usepackage{newfloat}
\usepackage{listings}
\usepackage{algorithm}
\usepackage{algorithmic}
\usepackage[dvipsnames]{xcolor}
\usepackage{arydshln} 
\usepackage{booktabs}
\usepackage{multirow}
\usepackage{amsmath}
\usepackage{amssymb}
\usepackage{colortbl}
\usepackage{tabularray}
\usepackage{bm}
\usepackage{subfigure}

\definecolor{colorFst}{HTML}{F59194}      % first
\definecolor{colorSnd}{HTML}{FAC791} % second
\definecolor{colorThd}{HTML}{FFFF99}   % third
\definecolor{url_color}{RGB}{42, 83, 163}

\DeclareCaptionStyle{ruled}{labelfont=normalfont,labelsep=colon,strut=off} % DO NOT CHANGE THIS
\floatstyle{ruled}
\newfloat{listing}{tb}{lst}{}
\floatname{listing}{Listing}
\title{Physics-Informed Deformable Gaussian Splatting: \\Towards Unified Constitutive Laws for Time-Evolving Material Field}

\author{
\normalsize{
    Haoqin Hong$^1$\equalcontrib},
    Ding Fan$^1$\equalcontrib,
    Fubin Dou$^1$,
    Zhili Zhou$^3$,
    Haoran Sun$^1$,
    Congcong Zhu$^{124}$\thanks{Corresponding authors.},
    Jingrun Chen$^{124}$\footnotemark[2]
    }

\affiliations{
    \textsuperscript{\rm 1} University of Science and Technology of China, China \\
    
    \textsuperscript{\rm 2} Suzhou Institute for Advanced Research, University of Science and Technology of China, China\\
    
    \textsuperscript{\rm 3} University of Illinois Urbana-Champaign, USA\\
    
    \textsuperscript{\rm 4} Suzhou Big Data \& AI Research and Engineering Center, China\\

}

\usepackage{bibentry}
\begin{document}

\maketitle

\begin{abstract}
Recently, 3D Gaussian Splatting (3DGS), an explicit scene representation technique, has shown significant promise for dynamic novel-view synthesis from monocular video input. However, purely data-driven 3DGS often struggles to capture the diverse physics-driven motion patterns in dynamic scenes. To fill this gap, we propose Physics‑Informed Deformable Gaussian Splatting (PIDG), which treats each Gaussian particle as a Lagrangian material point with time-varying constitutive parameters and is supervised by 2D optical flow via motion projection. Specifically, we adopt static-dynamic decoupled 4D decomposed hash encoding to reconstruct geometry and motion efficiently. Subsequently, we impose the Cauchy momentum residual as a physics constraint, enabling independent prediction of each particle’s velocity and constitutive stress via a time-evolving material field. Finally, we further supervise data fitting by matching Lagrangian particle flow to camera-compensated optical flow, which accelerates convergence and improves generalization. Experiments on a custom physics-driven dataset as well as on standard synthetic and real-world datasets demonstrate significant gains in physical consistency and monocular dynamic reconstruction quality.
\end{abstract}

\begin{links}
    \link{Code}{https://github.com/SCAILab-USTC/Physics-Informed-Deformable-Gaussian-Splatting}
\end{links}

% \begin{figure*}[t]
% \centering
% \includegraphics[width=2.1\columnwidth]{figs/insights - v2.pdf} % Reduce the figure size so that it is slightly narrower than the column. Don't use precise values for figure width.This setup will avoid overfull boxes.
% \caption{Insights}
% \label{insights}
% \end{figure*}

\begin{figure*}[ht]
\centering
\includegraphics[width=2\columnwidth]{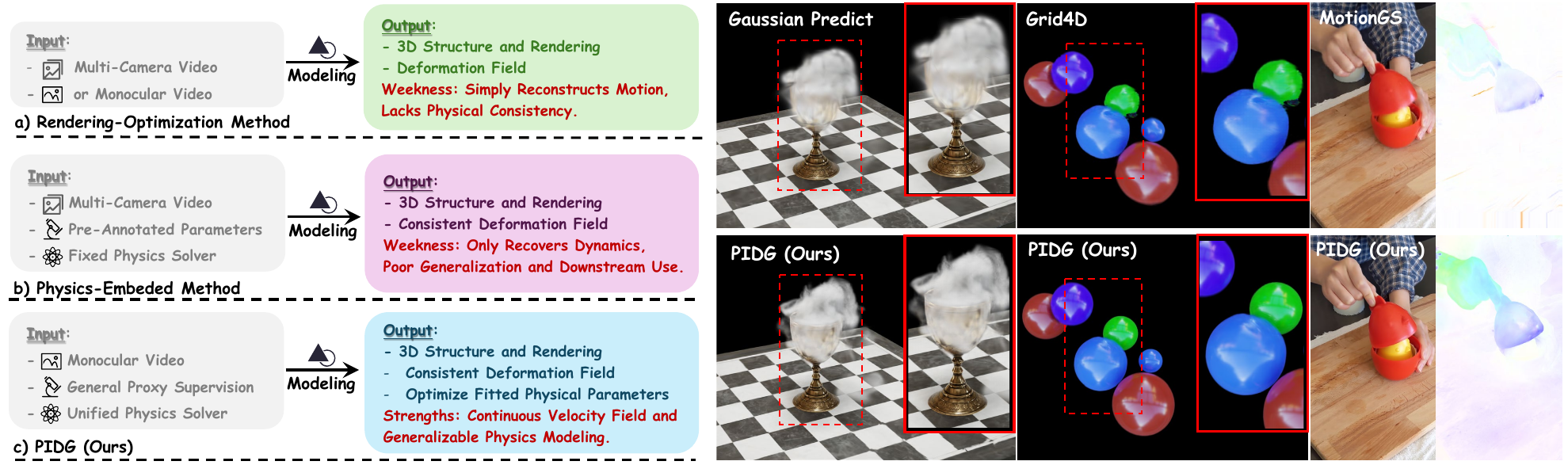} % Reduce the figure size so that it is slightly narrower than the column. Don't use precise values for figure width.This setup will avoid overfull boxes.
\caption{Physics-Informed Deformable Gaussian Splatting achieves physically consistent and generalisable monocular dynamic novel-view synthesis effects, performing excellently in physical scenarios such as fluid dynamics and elastic mechanics.}
\label{insights}
\end{figure*}

\section{Introduction}
\label{sec:Introduction}
% Can we achieve modelling of spatio-temporal structure and physical properties? In recent years, novel view synthesis techniques have shown great promise for creating digital representations of the physical world~\cite{yang2024thinkingspacemultimodallarge}. Their core objective is to generate images of a scene from any specified viewpoint in time, typically relying on multiple 2D input images to achieve high-fidelity 3D reconstruction~\cite{mildenhall2019local,schonberger2016structure, pumarola2021d}. We expect these methods not only to reconstruct known 3D geometry, but also to capture the spatio-temporal evolution of a scene’s intrinsic physical properties.

How can physical dynamics be faithfully embedded into spatio-temporal scene representations? Recent novel-view synthesis methods have demonstrated a remarkable ability to reconstruct high-fidelity 3D representations of the physical world from sparse 2D observations~\cite{yang2024thinkingspacemultimodallarge,mildenhall2019local,schoenberger2016sfm,pumarola2021d}.  
These approaches aim to render dynamic scenes from arbitrary viewpoints and timestamps, with the expectation of not only recovering dynamic geometry but also accurately capturing the evolution of intrinsic physical properties across space and time.

Due to the absence of credible constraints on motion pattern or 3D structure, novel-view synthesis in dynamic scenes faces significant challenges~\cite{zhao2024gaussianprediction,liu2025physgen,li20223d,xugrid4d,zhumotiongs}: firstly, sparse-view observations may be insufficient to supervise complex motion patterns; secondly, the synthesis results often suffer from inconsistencies of physical dynamics. To this end, we investigate how to embed motion constitutive laws into the dynamic novel view synthesis framework.

Neural Radiation Fields (NeRF) ~\cite{mildenhall2020nerf,pumarola2021d,park2021hypernerf} uses implicit scene representation and volume rendering to achieve novel view synthesis of high quality, but treats each reconstruction as a whole Eulerian volume, thus restricting the physical consistency to a two-dimensional projection. 3D Gaussian Splatting (3DGS) ~\cite{ kerbl20233d} enables real-time rendering with explicit Gaussian primitives, but makes it difficult to build 4D dynamic models from sparse inputs. The incremental method ~\cite{luiten2024dynamic} leads to an explosion in storage over time, whereas the deformation field method ~\cite{wu20244d,yang2024deformable,bae2024per} learns dynamic Gaussian motions using an implicit method. And the efficiency of dynamic reconstruction is further optimised using the low-rank assumption~\cite{huang2024sc, D-2DGS} or 4D decomposed hash coding~\cite{xugrid4d,4gvml,kwak2025modec}. Although these existing methods achieve considerable results in dynamic novel view synthesis, they only simplify the motion to rigid body transformation, ignoring the motion constitutive laws of different materials. Meanwhile, relying only on the visual supervision from monocular videos results in the inability to uniformly constrain the state of 3D particles, leading to deviations of Gaussian particles from their original motion patterns. 

To address these challenges, as shown in Fig.~\ref{insights}, we propose Physics-Informed Deformable Gaussian Splatting, which models dynamic 3D Gaussians as time-evolving Lagrangian particles to recover spatio-temporal structure and intrinsic physical properties. Our main contributions are:
  
\begin{itemize}
  \item We extend the Cauchy momentum equation into a constitutive law, in which each Gaussian is treated as a Lagrangian particle whose velocity and stress are predicted by a time-evolving material field, and momentum residuals are penalized to enforce physical consistency.
   \item We adopt 4D decomposed hash encoding by decoupling the spatio-temporal representation in a canonical hash space, significantly reducing memory from \(\mathcal{O}(n^4)\) to \(\mathcal{O}(n^3)\) while keeping satisfying reconstruction accuracy.
    \item We propose a Lagrangian particle flow matching term. By aligning Gaussian flow and velocity flow to camera-compensated optical flow, we directly supervise particle motion and ensure the material field converges reliably.
  % \item Our method is integrated into a fully differentiable pipeline that renders dynamic scenes from sparse monocular inputs, and demonstrates promising performance on our physics-driven dataset as well as on existing real-world and synthetic benchmarks.
  \item Our method integrates into a fully differentiable pipeline that renders dynamic scenes from sparse inputs, showing promising results on our physics-driven dataset and existing real-world and synthetic datasets.
\end{itemize}

\section{Related Work}
\label{sec:Related}
Early dynamic reconstruction methods based on explicit geometry~\cite{broxton2020immersive,newcombe2015dynamicfusion,orts2016holoportation,mildenhall2019local} and NeRF canonical mappings~\cite{mildenhall2020nerf,park2021nerfies,pumarola2021d} suffer from high computational costs. Recent 3D Gaussian Splatting approaches~\cite{kerbl20233d,luiten2024dynamic,wu20244d,xugrid4d,huang2024sc, D-2DGS} improve efficiency and detail but still struggle to generalize dynamic motion while preserving static accuracy. To address 2D visual supervision limitations, methods like GaussianFlow~\cite{gao2024gaussianflow}, MotionGS~\cite{zhumotiongs}, and MAGS~\cite{guo2024motion} incorporate optical flow, achieving high-quality rendering yet lacking physical consistency. 

Physics-guided methods using material point method (MPM)~\cite{zhang2025physdreamer,xie2024physgaussian}, physics-informed neural networks (PINNs)~\cite{chu2022physics,wang2024physics}, or spring-mass models~\cite{zhong2025reconstruction,jiang2025phystwinphysicsinformedreconstructionsimulation} enhance realism but remain limited by discretization, fixed material properties, and poor adaptability to diverse scenes, leaving the development of a universal, physically consistent framework for diverse dynamic scenes an open challenge.

\section{Preliminaries}
\label{sec:Preliminary}
\subsection{Physics-Informed Neural Network}
PINNs~\cite{raissi2019physics} establish a synergistic framework between data-driven modeling and physical law enforcement by encoding governing equations into loss functions. The core architecture employs partial differential equation (PDE) residuals as supervisory signals through a multi-task learning paradigm. 

% The total loss function combines data fitting and physical constraint terms: 

% \begin{equation}
%     \mathcal{L}_{\text{PINN}} = \underbrace{\mathcal{L}_{\text{Data}}}_{\text{Data Fitting Term}} + \underbrace{\mathcal{L}_{\text{PDE}}}_{\text{Physical Constraint Term}}.
% \end{equation}

The total loss function combines data fitting and physical constraint terms. The data fitting term quantifies prediction errors relative to measurements through mean squared error minimization, where $N_d$ is the number of training data points. The physical constraints are enforced through automatic differentiation for PDE residual computation across $N_f$ points within the physical domain $\Omega$:

{\small \begin{align}
    \mathcal{L}_{\text{DF}} &= \frac{1}{N_d} \sum_{i=1}^{N_d} \| u_{\text{NN}}(\mathbf{x}_i) - u_{\text{true}}(\mathbf{x}_i) \|^2, \label{data_fitting}\\
    \mathcal{L}_{\text{PDE}} &= \frac{1}{N_f} \sum_{i=1}^{N_f} \| 
\nabla_{\mathbf{x}} \cdot \mathbf{F}(u_{\text{NN}}, \mathbf{x}) \|_{\Omega}^2. \label{PDE}
\end{align}}

Here, \(u_{\mathrm{NN}}\) is the neural network’s predicted solution, \(u_{\mathrm{true}}\) is the ground-truth solution from measurements or analytic solution, \(\mathbf{F}\) is the flux in conservation-form PDEs, \(\nabla_{\mathbf{x}}\!\cdot\!\mathbf{F} = \sum_{i=1}^d \frac{\partial F_i}{\partial x_i}\) denotes its divergence in \(d\) spatial dimensions, and \(\mathbf{x}_i\in\Omega\) are the evaluation points.

% where $u_{\text{NN}}$ represents the neural network's predicted solution, $u_{\text{true}}$ denotes the ground truth solution from physical measurements or analytical solutions, $\mathbf{F}$ denotes the flux function in conservation-form PDEs, $\nabla_{\mathbf{x}} \cdot \mathbf{F} = \sum_{i=1}^d \frac{\partial F_i}{\partial x_i}$ represents the divergence of the flux field, with $d$ being the spatial dimension, and $\mathbf{x}_i \in \Omega$ corresponds to the spatial coordinates of evaluation points. 
% The core advantage of PINNs lies in integrating physical laws with data-driven modeling to enhance generalization capabilities for complex systems.

\subsection{3D Gaussian Splatting Method}
% 3DGS~\cite{kerbl20233d} represents a scene as an explicit set of millions of learnable 3D Gaussians:
% \begin{enumerate}
%     \item Center position $\mu \in \mathbb{R}^3$;
%     \item Orientation as a quaternion $\mathbf{q} \in \mathbb{R}^4$;
%     \item Anisotropic scale $\mathbf{s} \in \mathbb{R}^3$;
%     \item View-dependent color via spherical harmonics coefficients $\mathbf{h} \in \mathbb{R}^{3(k+1)^2}$, yielding RGB color $\mathbf{c} \in \mathbb{R}^3$;
%     \item Opacity $o \in [0,1]$.
% \end{enumerate}

3DGS~\cite{kerbl20233d} represents a scene as an explicit set of millions of learnable 3D Gaussians, each parameterized by a center position $\mu\in\mathbb{R}^3$, a quaternion orientation $q \in\mathbb{R}^4$, an anisotropic scale $s\in\mathbb{R}^3$, spherical harmonics coefficients $h \in\mathbb{R}^{3(k+1)^2}$ for view-dependent color $c\in\mathbb{R}^3$, and an opacity $o\in[0,1]$. Each Gaussian contributes at point \(\mathbf{x}\in\mathbb{R}^3\) as $\alpha_i(\mathbf{x}) = o_i \exp\bigl(-\tfrac12(\mathbf{x}-\mu_i)^\top \Sigma_i^{-1}\,(\mathbf{x}-\mu_i)\bigr)$, where \(\Sigma_i\) is computed from \(q_i\) and \(s_i\). This infinite‐support formulation enables global gradient flow.

% Each Gaussian softly occupies space, contributing to the scene at position $\mathbf{x} \in \mathbb{R}^3$ according to:
% \begin{equation}
%     \varphi_i(\mathbf{x}) = o_i \exp\left(-\frac{1}{2}(\mathbf{x} - \mu_i)^T \mathbf{\Sigma}_i^T (\mathbf{x} - \mu_i)\right),
% \label{eq:gaussian}
% \end{equation}
% where the covariance matrix $\mathbf{\Sigma}_i \in \mathbb{R}^{3 \times 3}$ is derived from $\mathbf{q}_i$ and $\mathbf{s}_i$. This formulation results in Gaussians with theoretically infinite spatial support, allowing global gradient influence during optimization.

Differentiable rendering in 3DGS depends on a splatting pipeline: each Gaussian is projected onto the image plane via its mean \(\mu_i\) and covariance \(\Sigma_i\), and contributes according to a 2D Gaussian. When a pixel is covered by \(N\) depth-sorted Gaussians, we composite its color as
$\mathbf{C} = \sum_{i=1}^N c_i\,\alpha_i T_{i}$, where \(c_i\) and \(T_i= \prod_{j=1}^{i-1}(1 - \alpha_j)\) are the RGB color and cumulative transmittance of the \(i\)-th Gaussian. The full scene is parameterized by
$P = \{\,G_i = (\mu_i, q_i, s_i, c_i, o_i)\,\}_{i=1}^N$.

\subsection{Problem Formulation}
\subsubsection{Known and Unknown.} 
Novel view synthesis in monocular dynamic scenes requires reconstructing complex spatio-temporal evolution from sparse video frames. Given a sequence of monocular inputs, the goal is to estimate a set of Gaussian particles with static attributes and a time-varying deformation field that models their motion over time.

% The task of novel view synthesis in monocular dynamic scenes faces the challenge of reconstructing the complex spatio-temporal evolution from sparse video frames. Given a sequence of sparse monocular frames 
% $\{I_t\}_{t=1}^T$. 
% We seek to estimate \(N\) Gaussian particles with static attributes $\{\mu_i^c,\;\Sigma_i,\;c_i\}_{i=1}^N$ and a time‐varying deformation field $F : (\mu_i^c,\,t)\;\mapsto\;\mu_i(t)$.

\subsubsection{Objective.}
The optimization objectives are twofold as rendering consistency and physical consistency. The former ensures that the synthesized views faithfully reproduce the input frames, while the latter imposes physics-driven constraints to encourage plausible, convergent, and generalizable dynamic behavior. The motivation for this work arises from the following key questions, which we attempt to investigate in this paper: (1) How can we model the positions and time-varying deformations of Gaussian particles without any known ground-truth particle motion priors? (2) What kind of boundary conditions or surrogate supervision can enable physically meaningful and generalizable modeling of dynamic materials?
% \begin{enumerate}
%  \item How can we model the positions and time-varying deformations of Gaussian particles without any known ground-truth particle motion priors?
%   \item What kind of boundary conditions or surrogate supervision can enable physically meaningful and generalizable modeling of dynamic materials?
% \end{enumerate}

\section{Method Overview}
\label{sec:Method}
\begin{figure*}[t]
\centering
\includegraphics[width=2.0\columnwidth]{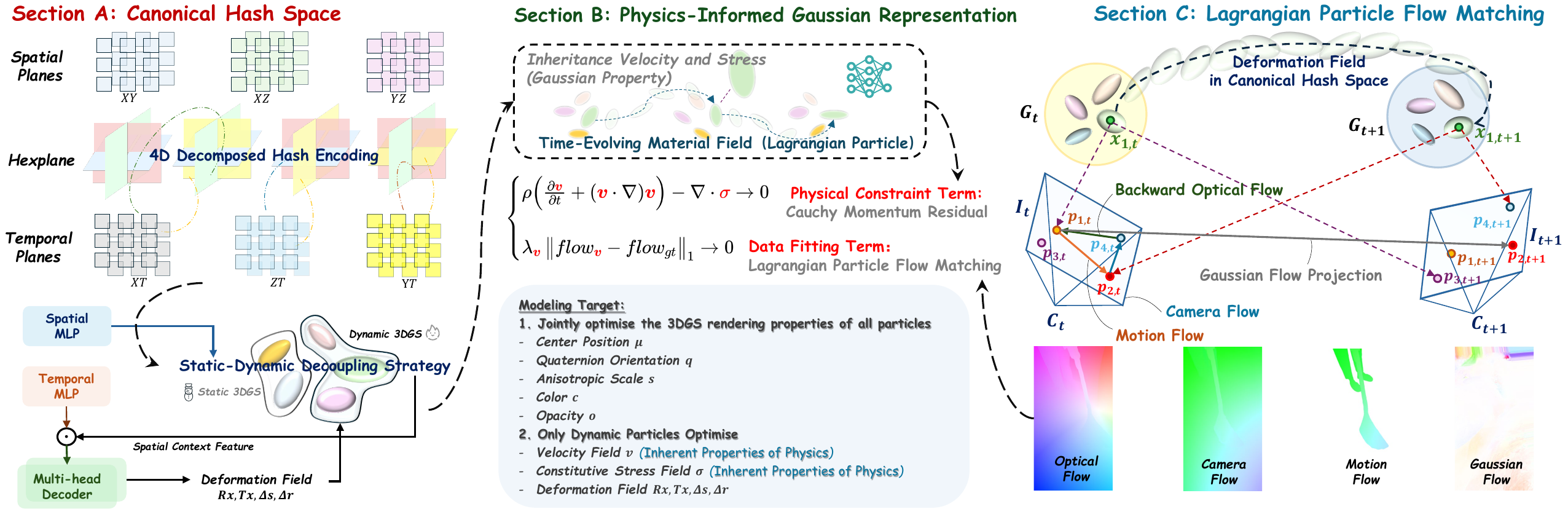} % Reduce the figure size so that it is slightly narrower than the column. Don't use precise values for figure width.This setup will avoid overfull boxes.
\caption{\textbf{Overview.} It integrates dynamic reconstruction in the canonical hash space (Sec.~\ref{sec:Method}.1), physics-informed Gaussian representation (Sec.~\ref{sec:Method}.2), and Lagrangian particle flow matching (Sec.~\ref{sec:Method}.3) to achieve differentiable and physically consistent monocular dynamic video reconstruction. Training architecture can be found in Supp. Sec.~C.}
\label{overview}
\end{figure*}

\subsection{Dynamic Modeling in Canonical Hash Space}
% Recent research commonly adopts a strategy to reconstruct monocular dynamic scenes by establishing a canonical space where geometry remains static, and then learns deformation fields to warp this 3D representation to each time step; however, incremental or deformation-field-based methods ~\cite{luiten2024dynamic,yang2024deformable} face significant memory overhead, while low-rank plane decomposition methods ~\cite{huang2024sc} often degrade motion fidelity. Moreover, because dynamic regions are inherently sparse, developing mechanisms to robustly perceive and accurately model these sparse, time-varying elements is equally critical. 
Recent methods reconstruct dynamic scenes by defining a static canonical space and learning deformation fields for temporal mapping. However, incremental and deformation-field approaches~\cite{luiten2024dynamic,yang2024deformable} incur high memory overhead, while low-rank plane decomposition~\cite{huang2024sc} often reduces motion fidelity. The sparsity of dynamic regions further underscores the need to perceive and model time-varying elements robustly.

\subsubsection{4D Decomposed Hash Encoding.}
Inspired by hash-based Grid4D~\cite{xugrid4d}, we replace costly 4D MLPs or low-rank plane decompositions with a compact hash encoding,  which reduces memory from \(\mathcal{O}(n^4)\) to \(\mathcal{O}(n^3)\), where \(n\) is samples per axis. Specifically, each Gaussian coordinate \((x,y,z,t)\) is mapped to four separate 3D hash grids:
{\small \begin{equation}
    G_{xyz}(x,y,z),  
    G_{xyt}(x,y,t), 
    G_{yzt}(y,z,t),  
    G_{xzt}(x,z,t),
\end{equation}}

We then use a Spatial MLP \(f_s\) to extract a spatial context feature and convert it into a directional attention weight:
{\small \begin{equation}
    a \;=\; 2\,\sigma\!\bigl(f_s(G_{xyz}(x,y,z))\bigr) - 1, \quad a\in(-1,1),
\end{equation}}
where \(\sigma(\cdot)\) denotes the Sigmoid function. The weight of the spatial context feature modulates the raw temporal encoding produced by Temporal MLP \(f_t\) so that motion predictions respect both spatial structure and temporal variation:
{\small \begin{equation}
    h \;=\; a \;\odot\;
    f_t\bigl(
      G_{xyt}(x,y,t),\,G_{yzt}(y,z,t),\,G_{xzt}(x,z,t)
    \bigr).
\end{equation} }

% Finally, to track Gaussian deformation, we adopt a compact multi‐head MLP \(D(h) = \{R_x,\;T_x,\;\Delta r,\;\Delta s\}\) to map the feature \(h\) to a translation \(T_x\), a rotation increment based on quaternions \(\Delta r\), and a scale update \(\Delta s\). We convert \(\Delta r\) into a rotation matrix \(R_x\) and update the canonical Gaussian parameters:

To track deformation, we finally use a compact multi-head MLP \(D(h) = \{R_x,\;T_x,\;\Delta r,\;\Delta s\}\) that decodes feature \(h\) into a rotation \(R_x\) and translation \(T_x\), a scale update \(\Delta s\), and a rotation increment \(\Delta r\) represented as a quaternion. The quaternion \(\Delta r\) is converted to an incremental rotation matrix \(\Delta R\) for smooth interpolation and stable optimization. The canonical Gaussian is updated as:
{\small \begin{equation}
  \mu' = R_x\,\mu + T_x, 
  S'   = S + \Delta s, 
  R'   = R + \Delta r.
\end{equation}}

% \begin{equation}
%   D(h) = \{R_x,\;T_x,\;\Delta r,\;\Delta s\}.
% \end{equation}

\subsubsection{Decoupling and Canonical Hash Space.}
4D decomposed hash encoding enables dynamic reconstruction with differentiable splatting, where all rendering parameters are optimized using the renders loss in 3DGS~\cite{kerbl20233d}:
{\small \begin{equation}
    \mathcal{L}_{\rm renders} = (1 - \lambda_c)\,\mathcal{L}_1 + \lambda_c\,\mathcal{L}_{\rm D-SSIM}\,.  
\end{equation}}

However, only one of its four 3D grids \(G_{xyz}\) is used to encode static geometry, limiting the capacity for static representation. 
To overcome this, we adopt a two-stage optimization strategy. 
First, all Gaussians are densified to jointly optimize static geometry and dynamic deformation. 
Next, densification is paused, and a dynamic mask is applied to isolate moving regions. 
During refinement, masked Gaussians update their deformation fields, whereas unmasked ones remain frozen, only their rendering parameters are fine-tuned within the canonical hash space. 
This procedure yields an efficient separation between static and dynamic components, enabling subsequent physical modeling of dynamics while preserving static representations to the greatest extent.

% However, only one of its four 3D grids \(G_{xyz}\) is used to encode static geometry, which limits the model’s capacity for static representation. To address this, we adopt a two-stage approach: first, densify all Gaussians and jointly optimize static geometry and dynamic deformation; then, pause densification and apply a dynamic mask to isolate moving regions. During refinement, masked Gaussians update deformation fields, while unmasked ones are partially frozen, only their canonical hash space parameters are fine-tuned, with all other attributes fixed. This achieves a clear and efficient representation of static-dynamic separation.

% 4D decomposed hash encoding enables dynamic reconstruction with differentiable splatting~\cite{kerbl20233d}, where all rendering parameters are optimized using the Gaussian Splatting loss:
% \begin{equation}
%   L = (1 - \lambda_c)\,L_1 + \lambda_c\,L_{D\text{-}SSIM}.
% \end{equation}
% However, only one of its four 3D grids $G_{xyz}$ encodes static geometry, limiting static capacity. To address this, after joint optimization we halt densification updates and apply SamV2~\cite{ravi2024sam} to mask dynamic regions. During refinement, only Gaussians outside the mask update their deformation fields, all Gaussians continue optimizing splatting parameters in the canonical hash space, and the spatial MLP $f_s$ is frozen to preserve the static structure, yielding a static and dynamic decoupled canonical hash space.

\subsection{Physics-Informed Gaussian Representation}
Although 4D decomposed hash encoding enables efficient dynamic reconstruction, it typically overlooks the intrinsic motion pattern, impairing physical coherence. Hence, we treat each Gaussian as a Lagrangian particle within the canonical hash space and explicitly embed continuum mechanics constitutive laws into this framework. Consequently, particle trajectories and internal states evolve naturally, maintaining consistent physical properties and coupling rendering attributes seamlessly with physical behaviors, eliminating the costly rematching after densification.

% 4D Decomposed Hash Encoding offers a favourable balance between computational efficiency and dynamic reconstruction capability. However, they frequently ignore the inherent motion properties of objects, resulting in a lack of physical coherence. To further improve the motion modelling capabilities, we treat each Gaussian as a Lagrangian particle in canonical hash space, which allows for the unification of the motion constitutive laws in the framework of continuum mechanics. The trajectories and internal states of the dynamic Gaussian particles evolve, thus preserving the characteristics of each particle and directly modelling the coupling between motion and stress without the need for costly re-matching after densification.

\subsubsection{Constitutive Laws in Continuum Mechanics.}
In continuum mechanics, motion follows from the conservation of mass, momentum, and energy~\cite{landau1987fluid, nicolis2015zoology}. For a control volume \(V\) with density \(\rho\) and velocity field \(\boldsymbol{v}\), Newton’s second law gives
{\small \begin{equation}
    \frac{\mathrm{d}}{\mathrm{d}t}\int_V \rho\,\boldsymbol{v}\,\mathrm{d}V
= \int_V \mathbf{f}\,\mathrm{d}V
+ \oint_{\partial V} \boldsymbol{\sigma}\cdot\mathbf{n}\,\mathrm{d}S,
\end{equation}} where \(\mathbf{f}\) is the body‐force density and \(\boldsymbol{\sigma}\) is the Cauchy stress tensor. Applying the divergence theorem $\oint_{\partial V} \boldsymbol{\sigma}\cdot\mathbf{n}\,\mathrm{d}S
= \int_V (\nabla\cdot\boldsymbol{\sigma})\,\mathrm{d}V$
yields the differential form

{\small \begin{equation}
    \rho\biggl(\frac{\partial\boldsymbol{v}}{\partial t} + \boldsymbol{v}\cdot\nabla\boldsymbol{v}\biggr)
= \nabla\cdot\boldsymbol{\sigma} + \mathbf{f}.
\label{con:cme}
\end{equation}}

The Cauchy momentum equation generalizes \(m\mathbf{a}=F\) to continua by equating the material derivative of momentum to body and surface forces. In the Lagrangian frame, the stress tensor decomposes as \(\boldsymbol{\sigma} = \boldsymbol{\sigma}_{\mathrm{aniso}} + \mu\bigl(\boldsymbol{\varepsilon} - \tfrac{1}{3}\mathrm{tr}(\boldsymbol{\varepsilon})\,\mathbf{I}\bigr)\), where \(\mu\) is the shear viscosity and \(\mathbf{I}\) the identity, and its divergence $\nabla\cdot\boldsymbol{\sigma} = \sum_j \frac{\partial \sigma_{ij}}{\partial x_j}$ quantifies the internal force density balancing inertia. By choosing different constitutive laws for \(\boldsymbol{\sigma}\), it reduces to the governing equations for fluids, elastic solids, or rigid bodies~\cite{crossley2017effectivefieldtheorydissipative,Glorioso:2017fpd}. Details of evolvability can be found in Supplementary Sec.~A.

% Detailed derivations of these reductions can be found in Supplementary Material Sec.~A.

% This Cauchy momentum equation generalizes \(m\mathbf{a}=\mathbf{F}\) to continua, the rate of momentum change equals body and surface forces. From continuum mechanics, the stress tensor in the Lagrangian perspective decomposes into an anisotropic part and a viscous correction proportional to the strain‐rate tensor \(\boldsymbol{\varepsilon}\): $\boldsymbol{\sigma} = \boldsymbol{\sigma}_{\text{aniso}}
%          + \mu \Bigl(\boldsymbol{\varepsilon}
%          - \tfrac{1}{3}\mathrm{tr}(\boldsymbol{\varepsilon})\,\mathbf{I}\Bigr)$, where \(\mu\) is the shear viscosity and \(\mathbf{I}\) the identity. Its divergence, $\nabla\cdot\boldsymbol{\sigma} = \sum_j \frac{\partial \sigma_{ij}}{\partial x_j}$, quantifies the net internal force density that must balance inertial effects. By choosing different constitutive laws for \(\boldsymbol{\sigma}\), it reduces to the governing equations for fluids, elastic solids, or rigid bodies. Detailed derivations of these reductions can be found in Supplementary Material Section~A.
% \begin{equation}
%   \boldsymbol{\sigma} = \boldsymbol{\sigma}_{\text{aniso}}
%          + \mu \Bigl(\boldsymbol{\varepsilon}
%          - \tfrac{1}{3}\mathrm{tr}(\boldsymbol{\varepsilon})\,\mathbf{I}\Bigr),
% \end{equation}

\subsubsection{Time-Evolving Material Field.}

% As shown in Fig.~\ref{overview} Sec.~B, building on the canonical hash space, we further embed the normalized 4D coordinate \(x=(x,y,z,t)\) into six learnable tensors $\mathbf{F}_{\rm Hash} \in\mathbb{R}^{6} $ to obtain expressive, yet efficient features, including spatial planes $(XZ, XY, YZ)$ and temporal planes $(XT, YT, ZT)$. These planes capture local correlations along different axis‐aligned slices, preserving fine-grained structure. We further append a Fourier time encoding $T(t) = \bigl[\sin(\omega_1 t),\cos(\omega_1 t),\dots,\sin(\omega_n t),\cos(\omega_n t)\bigr] \in\mathbb{R}^{2n}$ to model high-frequency temporal variation, and include a learnable index embedding $\mathbf{e}_{i=1,2,\dots,N}\in\mathbb{R}^{H}$, where \(N\) is the total number of Gaussian particles and \(H\) is the embedding dimension. The feature vector is $\mathbf{F}
% = \bigl[\mathbf{F}_{\rm Hash},\;T(t),\;\mathbf{e}_i\bigr]\;\in\mathbb{R}^{6 + 2n + H}$.

Building on the canonical hash space, as shown in Fig.~\ref{overview} Sec.~B, we embed the normalized 4D coordinate \(x=(x,y,z,t)\) into six learnable tensors \(\mathbf{F}_{\rm Hash}\in\mathbb{R}^6\) representing the spatial planes $XZ, XY, YZ$ and the temporal planes $XT, YT, ZT$ to capture axis-aligned correlations and preserve fine-grained structure. We further append a Fourier time encoding 
$T(t)=\bigl[\sin(\omega_1 t),\cos(\omega_1 t),\dots,\sin(\omega_n t),\cos(\omega_n t)\bigr]\in\mathbb{R}^{2n}$
for high-frequency temporal variation, and include a learnable index embedding \(\mathbf{e}_{i=1,\dots,N}\in\mathbb{R}^{H}\), where \(N\) is the number of Gaussian particles and \(H\) is the embedding dimension, to encode each particle's intrinsic physical attributes, yielding the feature vector
$\mathbf{F}=\bigl[\mathbf{F}_{\rm Hash},\;T(t),\;\mathbf{e}_i\bigr]\in\mathbb{R}^{6+2n+H}$.

A multi‐head MLP \(f_\theta\) ingests   \(\mathbf{F}\) and jointly predicts the velocity vector $  \boldsymbol{v}(x,t) = \bigl(v_x(x,t),\,v_y(x,t),\,v_z(x,t)\bigr)\in\mathbb{R}^{3}$
and the six independent components of the stress tensor $  \boldsymbol{\sigma}(x,t) = \bigl(\sigma_{xx},\,\sigma_{yy},\,\sigma_{zz},\,\sigma_{xy},\,\sigma_{xz},\,\sigma_{yz}\bigr)\in\mathbb{R}^{6}$. These outputs of our time-evolving material field are compactly written as $\bigl(\boldsymbol{v},\,\boldsymbol{\sigma}\bigr)
  = f_\theta\bigl(\mathbf{F}\bigr)$.
% \begin{equation}
%   \bigl(\boldsymbol{v},\,\boldsymbol{\sigma}\bigr)
%   = f_\theta\bigl(\mathbf{F}\bigr).
% \end{equation}

% Thus, as shown in Fig.~\ref{overview} Sec. B, we generalize the local Cauchy momentum balance of Eq. (\ref{con:cme})
% into the global physics constraint term defined in Eq. (\ref{PDE}) by introducing the residual $
%   \mathbf{r}(x,t)
%   = \rho\Bigl(\frac{\partial\boldsymbol{v}}{\partial t} + (\boldsymbol{v}\cdot\nabla)\boldsymbol{v}\Bigr)
%     - \nabla\cdot\sigma$, and penalize its \(L_2\) norm over \(M\) samples to obtain
% \begin{equation}
%   \mathcal{L}_{\rm CMR}
%   = \frac{1}{M}\sum_{i=1}^{M}\|\mathbf{r}(x_i,t_i)\|_2^2 \,,
% \end{equation} and the physics-informed material field integrated end-to-end via $\mathcal{L}_{\rm CMR}$, enables fully differentiable, physics‑consistent prediction of each Gaussian’s temporal evolution of velocity and constitutive stress, while ensuring satisfactory reconstruction quality. In this formulation, each Gaussian $G_i = (\mu_i,\,q_i,\,s_i,\,c_i,\,o_i,\,\mathbf{e}_i,\,\boldsymbol{v}_i,\,\boldsymbol{\sigma}_i)$ treats the velocity \(\boldsymbol{v}_i\) and constitutive stress \(\boldsymbol{\sigma}_i\) as independent intrinsic attributes.

Thus, we generalize the local Cauchy momentum balance of Eq.~(\ref{con:cme}) into the global physics constraint \(\mathcal{L}_{\rm PDE}\) of Eq.~(\ref{PDE}) by introducing the residual, omitting \(\mathbf{f}\) as its effect can be absorbed into the network-predicted velocity or stress:

{\small \begin{equation}
    \mathbf{r}(x,t)
  = \rho\Bigl(\frac{\partial\boldsymbol{v}}{\partial t} + (\boldsymbol{v}\cdot\nabla)\boldsymbol{v}\Bigr)
    - \nabla\cdot \boldsymbol{\sigma}
\end{equation}}
and penalizing its \(L_2\) norm over \(M\) samples to obtain
{\small \begin{equation}
      \mathcal{L}_{\rm CMR}
  = \frac{1}{M}\sum_{i=1}^{M}\|\mathbf{r}(x_i,t_i)\|_2^2,
\end{equation}} treating velocity \(\boldsymbol{v}_i\) and constitutive stress \(\boldsymbol{\sigma}_i\) as independent intrinsic attributes. This design allows each Gaussian particle to not only be encoded within the deformation field, but also to evolve continuously over time, updating its motion and internal states in a physically consistent manner.

\subsection{Lagrangian Particle Flow Matching}

\begin{figure}[H]
\centering
\includegraphics[width=0.9\columnwidth]{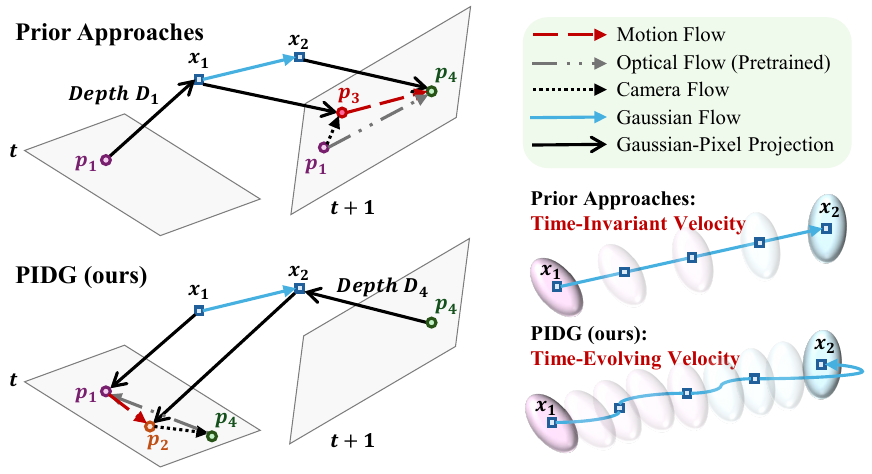} % Reduce the figure size so that it is slightly narrower than the column. Don't use precise values for figure width.This setup will avoid overfull boxes.
\caption{Forward vs Backward optical flow decomposition.}
\label{LPFM}
\end{figure}

Explicitly learning per-Gaussian velocity fields is essential for accurately capturing motion in dynamic novel-view synthesis. Prior works~\cite{4gvml} either treat velocity as a static attribute of each Gaussian or derive it solely from four-dimensional Gaussian slices, both assume time-invariant velocity and neglect temporal variation. Incorporating a physics-informed Gaussian representation effectively models independent particle velocities. However, enforcing only the Cauchy momentum residual often prevents velocity and stress predictions from converging to physically meaningful solutions. To this end, we propose Lagrangian particle flow matching: regarding the decomposed motion flow as an empirical measurement and incorporating it as a data-fitting term to directly supervise the predicted velocity field. By minimizing the discrepancy between rendered and observed motion flow, our method promotes convergence of both velocity and stress to physically plausible values.

\subsubsection{Optical Flow Decomposition.} 
To accurately capture the real motion of target objects, we follow a similar motion decomposition strategy ~\cite{zhumotiongs,you2025gavs}, explicitly decomposing optical flow between frame $I_{t}$ and $I_{t+1}$ into camera flow and motion flow, and treating motion flow as pseudo grounding truth to supervise the rendered flow of Gaussians and velocity field.  Notably, flow-based supervision must register $I_t$ and $I_{t+1}$ by warping one flow to the other. Forward schemes~\cite{zhumotiongs} mask the motion flow, and warp the observed Gaussian flow from $I_{t+1}$ to $I_{t}$, producing stripe artifacts in Gaussian flow. Our key observation is that, instead of directly masking stripe artifacts in Gaussian flow, we can compute the backward motion flow in $I_{t+1}$, warp it to $I_t$, apply motion masking there, and retain the original unwarped Gaussian flow in $I_t$. Since the mask operates on a pixel-wise field, stripe artifacts are effectively removed without corrupting true displacements, resulting in a clean supervisory signal.

As shown in Fig.~\ref{LPFM}, given two frames $I_t$ and $I_{t+1}$, the forward strategy~\cite{zhumotiongs,you2025gavs}  selects a pixel $p_1\!\in\!I_t$ with depth $D_1$ and camera intrinsic matrix $K$, back-projects it to 3D as $x_1 = K^{-1}D_1p_1$, and re-projects $x_1$ into $I_{t+1}$ to obtain $p_3$.  With the pretrained optical flow network~\cite{xu2023unifying} predicted displacement $flow^{f}_{t\to t+1} = p_4-p_1$ , the camera flow and motion flow are defined as $p_4 - p_1$ and $p_4 - p_3$, respectively. Our proposed backward motion flow computation starts from obtaining pixel $p_4\!\in\!I_{t+1}$ and its depth $D_4$, then back-projects to $x_2 = K^{-1}D_4p_4$ and re-projects $x_2$ into $I_t$ at $p_2$.  With the backward optical flow estimation $flow^{b}_{t+1 \to t} =  p_1 - p_4$, we define the backward camera flow and motion flow as $p_4 - p_2$ and $p_2 - p_1$, with ground‑truth flow as $flow_{gt} = p_2 - p_1$. Crucially, $flow_{gt}$ is already expressed in the reference frame $I_t$, allowing artifact-free supervision with motion mask.

\subsubsection{Gaussian Flow and Velocity Flow Matching.} 
Building on our physics-informed Gaussian representation, we follow GaussianFlow~\cite{gao2024gaussianflow}, compute a Lagrangian particle flow at each pixel \(p_1\) by tracking the top-\(K\) Gaussians, ranked by their splatting weights $w_{i} = \frac{\alpha_{i} T_{i}}{\sum_{j}\alpha_{j} T_{j}}$. Let \(\mu_{i,t}\) and \(\Sigma_{i,t}=U_i\Lambda_iV_i^\top\) be each Gaussian’s 2D center and covariance with singular value decomposition. Propagating Gaussians from time \(t\) to \(t+1\) according to its learned position update yields a displacement per particle, and summing the weighted displacements gives the Gaussian flow
{\small \begin{align}
    p^{\rm gaussian}_{i,t+1}
= \Lambda_{i+1}^{\frac12}\Lambda_{i}^{-\frac12}(p_1 - \mu_{i,t})
+ \mu_{i,t+1},\\
 flow_{g}
= \sum_{i=1}^{K} w_i\,\bigl(p^{\rm gaussian}_{i,t+1} - p_1\bigr).
\end{align}}
% \begin{equation}
%     p^{\rm gaussian}_{i,t+1}
% = \Lambda_{i+1}^{\frac12}\Lambda_{i}^{-\frac12}(p_1 - \mu_{i,t})
% + \mu_{i,t+1},
% \end{equation}

% \begin{equation}
%     flow_{g}
% = \sum_{i=1}^{K} w_i\,\bigl(p^{\rm gaussian}_{i,t+1} - p_1\bigr).
% \end{equation}

In parallel, we advect each Gaussian by its predicted velocity and project it to $I_{t}$, yielding \(v_t \in \mathbb{R}^{2}\) for velocity flow
{\small \begin{align}
    p^{\rm velocity}_{i,t+1}
= \Lambda_{i+1}^{\frac12}\Lambda_{i}^{-\frac12}(p_1 - \mu_{i,t})
+ v_t\,\Delta t + \mu_{i,t},\\
        flow_{v}
= \sum_{i=1}^{K} w_i\,\bigl(p^{\rm velocity}_{i,t+1} - p_1\bigr).
\end{align}}
Details of the calculation can be found in the Supp. Sec.~B. Finally, we treat the Lagrangian particle flow matching loss \(\mathcal{L}_{\rm LPFM}\) as a data-fitting term \(\mathcal{L}_{\rm DF}\) as Eq. (\ref{data_fitting}), which helps the predicted velocity of the time-evolving material field converge more rapidly.  The loss is defined as
{\small \begin{equation}
    \mathcal{L}_{\rm LPFM}
    = \lambda_g \,\bigl\lVert flow_{g} - flow_{gt}\bigr\rVert_{1}
    + \lambda_v \,\bigl\lVert flow_{v} - flow_{gt}\bigr\rVert_{1},
\end{equation}}
where $\lambda_g =\lambda_v=0.5$ are the flow-matching weights for Gaussian and velocity flow, and \(flow_{gt}\) denotes the motion flow with motion mask. By anchoring each particle’s Lagrangian trajectory to its physical state, this data-fitting term facilitates convergence of both velocity and stress fields to physically consistent solutions. Unlike boundary-condition-driven constraints, this surrogate supervision provides a more generalizable and physically grounded regularization.

\section{Experiments}
\label{sec:Exp}
\begin{table*}[htbp]
\centering
\small{
\begin{tabular*}{\textwidth}{@{\extracolsep{\fill}} l *{9}{c}@{}}
\toprule
\multirow{3}{*}{Method}
  & \multicolumn{3}{c}{Motion Kuro}
  & \multicolumn{3}{c}{Mechanics Cloth}
  & \multicolumn{3}{c}{Dry Ice} \\
  & \multicolumn{3}{c}{- Human Motion -}
  & \multicolumn{3}{c}{- Cloth Simulation -}
  & \multicolumn{3}{c}{- Fluid Smoke -} \\
  & PSNR $\uparrow$ & SSIM $\uparrow$ & LPIPS $\downarrow$
  & PSNR $\uparrow$ & SSIM $\uparrow$ & LPIPS $\downarrow$
  & PSNR $\uparrow$ & SSIM $\uparrow$ & LPIPS $\downarrow$ \\
\cmidrule(lr){2-4} \cmidrule(lr){5-7} \cmidrule(lr){8-10}
D-NeRF  ~\cite{pumarola2021d}
  & 22.80 & 0.872 & 0.121
  & 31.24 & 0.927 & 0.090
  & 17.74 & 0.698 & 0.223 \\
D-3DGS  ~\cite{yang2024deformable}
  & 26.61 & 0.949 & 0.053
  & 39.50 & 0.986 & \underline{0.033}
  & 24.26 & 0.879 & 0.163 \\
GaussianPredict ~\cite{zhao2024gaussianprediction}
  & 26.41 & \underline{0.952} & \underline{0.049}
  & 36.45 & 0.985 & 0.034
  & \textbf{27.23} & 0.901 & 0.143 \\
SC-GS ~\cite{huang2024sc}
  & 26.37 & 0.931 & \textbf{0.048}
  & 39.13 & \textbf{0.991} & 0.066
  & 23.59 & 0.861 & 0.154 \\
Grid4D ~\cite{xugrid4d}
  & \underline{26.89} & 0.951 & 0.052
  & 39.01 & 0.985 & 0.037
  & 25.33 & 0.896 & 0.149 \\
MoDec-GS ~\cite{kwak2025modec} & 23.16 & 0.941 & 0.054 & 22.89 & 0.905 & 0.114 & 25.95 & \underline{0.919} & 0.139 \\
D-2DGS ~\cite{D-2DGS} 
  & 25.99 & 0.938 & 0.053
  & \textbf{41.24} & \underline{0.989} & \textbf{0.023}
  & 23.79 & 0.863 & \textbf{0.131}\\

\midrule
PIDG ($w/o\ (\mathcal{L}_{\rm LPFM} + \mathcal{L}_{\rm CMR})$)  & 26.86 & 0.951 & 0.053
  & 39.23 & 0.986 & 0.036
  & 25.34 & 0.896 & 0.148\\
PIDG ($w/o\ \mathcal{L}_{\rm LPFM}$)
  & 26.85 & \underline{0.952} & 0.051 & \underline{39.97} & 0.985 & 0.035 & 25.55 & 0.897 & 0.148 \\
PIDG $\mathit{(Ours \ Full)}$
  & \textbf{26.97} & \textbf{0.953}  & 0.051
  & 39.96     & \textbf{0.991}     & 0.034
  & \underline{26.12}     & \textbf{0.926}     & \underline{0.138} \\
\midrule
\multirow{3}{*}{Method}
  & \multicolumn{3}{c}{Rubber Duck}
  & \multicolumn{3}{c}{Balls Reaction}
  & \multicolumn{3}{c}{Average} \\
  & \multicolumn{3}{c}{- Soft Body -}
  & \multicolumn{3}{c}{- Elastic Mechanics -}
  & \multicolumn{3}{c}{- Physical Scenes -} \\
  & PSNR $\uparrow$ & SSIM $\uparrow$ & LPIPS $\downarrow$
  & PSNR $\uparrow$ & SSIM $\uparrow$ & LPIPS $\downarrow$
  & PSNR $\uparrow$ & SSIM $\uparrow$ & LPIPS $\downarrow$ \\
\cmidrule(lr){2-4} \cmidrule(lr){5-7} \cmidrule(lr){8-10}
D-NeRF  ~\cite{pumarola2021d}
  & 20.73 & 0.915 & 0.145
  & 24.76 & 0.916 & 0.093
  & 23.45 & 0.866 & 0.124 \\
D-3DGS  ~\cite{yang2024deformable}
  & 26.27 & 0.967 & 0.047
  & 31.08 & 0.975 & 0.032
  & 29.54 & 0.951 & 0.066 \\
GaussianPredict ~\cite{zhao2024gaussianprediction}
  & 27.22 & 0.968 & 0.046
  & \textbf{33.53} & 0.973 & 0.040
  & 30.17 & \underline{0.957} & 0.062 \\
SC-GS ~\cite{huang2024sc}
  & 25.17 & 0.946 & 0.049
  & 29.72 & 0.969 & 0.034
  & 28.80 & 0.940 & 0.070 \\
Grid4D ~\cite{xugrid4d}
  & 27.83 & 0.970 & \underline{0.032}
  & 32.53 & 0.977 & 0.033
  & 30.32 & 0.956 & 0.061  \\
MoDec-GS ~\cite{kwak2025modec} & 27.27 & 0.971 & 0.033 & 31.05 & \underline{0.982} & 0.037\ & 26.06 & 0.944 & 0.075 \\
D-2DGS ~\cite{D-2DGS}
  & 26.20 & 0.961 & 0.057
  & 28.93 & 0.967 & 0.040
  & 29.23 & 0.944 & 0.061 \\
\midrule
PIDG ($w/o\ (\mathcal{L}_{\rm LPFM} + \mathcal{L}_{\rm CMR})$) & 28.07 & 0.971 & \underline{0.032}
  & 32.79 & 0.978 & \textbf{0.029}
  & 30.46  & 0.956 & 0.060 \\
PIDG ($w/o\ \mathcal{L}_{\rm LPFM}$)
  & \underline{28.34} & \underline{0.972} & \textbf{0.031} &  33.18  & 0.980 & \textbf{0.029} & \underline{30.78} & \underline{0.957} & \underline{0.059} \\
PIDG $\mathit{(Ours \ Full)}$
  & \textbf{28.43} & \textbf{0.976} & 0.037
  & \underline{33.31} & \textbf{0.988} & \underline{0.031}
  & \textbf{30.96} & \textbf{0.967}  & \textbf{0.058}\\
\bottomrule
\end{tabular*}}
% \caption{\textbf{Quantitative dynamic novel view synthesis results on the proposed PIDG dataset.} We also evaluate the PIDG method to ablate the effects of the time-evolving material field with $\mathcal{L}_{\rm CMR}$ and the Lagrangian particle flow matching $\mathcal{L}_{\rm LPFM}$.}
\caption{\textbf{Quantitative dynamic novel view synthesis results on the proposed PIDG dataset.}}
\label{exp:dynamic_nvs_pidg}
\end{table*}

\subsection{Experimental Setup}

\subsubsection{Datasets.} 

To evaluate our method across various dynamic scenes, we compare it against state-of-the-art methods on the D-NeRF synthetic dataset~\cite{pumarola2021d} and the HyperNeRF real-world dataset~\cite{park2021hypernerf}. We also construct a custom dataset, PIDG, using Blender and a physics solver to evaluate the model in complex dynamic scenes. It covers diverse physical scenarios, including human motion, cloth simulation, fluid smoke, soft-body dynamics, and elastic mechanics, enabling a comprehensive assessment of motion reconstruction. For real-world data, we use COLMAP~\cite{schoenberger2016sfm} estimates for camera poses and point clouds; for synthetic data, $100k$ random points and precise poses from BlenderNeRF~\cite{Raafat_BlenderNeRF_2024}. In the HyperNeRF real-world dataset~\cite{park2021hypernerf}, we employ pretrained UniMatch~\cite{xu2023unifying}, Distill Any Depth~\cite{he2025distill}, and SAMv2~\cite{ravi2024sam} to obtain optical flow, depth maps, and motion masks. 

% Due to their poor generalization to synthetic scenes, we instead export masks from Blender and compute optical flow with the Dual TV-L1~\cite{zach2007duality}. 

% \subsubsection{Preprocess.} 
% In the HyperNeRF real-world dataset~\cite{park2021hypernerf}, we use pretrained UniMatch~\cite{xu2023unifying}, Distill Any Depth~\cite{he2025distill}, and SAMv2~\cite{ravi2024sam} to obtain optical flow, depth maps, and motion masks. Due to their poor generalization to synthetic scenes, we instead export depth and mask images from Blender and compute forward/backward optical flow with OpenCV Dual TV-L1 algorithm~\cite{zach2007duality}. 

\subsubsection{Hyperparameters and Metrics.} 
Following Grid4D~\cite{xugrid4d}, we use a 16-level spatial hash encoder with resolutions from 16 to 2048 and set the temporal encoder’s maximum level $L$ to 32. Loss weights $\lambda_c$, $\lambda_{\rm{CMR}}$, and $\lambda_{\rm{LPFM}}$ are 0.2, 0.1, and 0.01, respectively. All methods are trained for $50 k$ iterations on synthetic and $40k$ on real-world datasets, with rendering resolutions of 800 × 800 for D-NeRF, 536 × 900 for HyperNeRF, and 1600 × 900 for PIDG. Experiments run on an 80GB NVIDIA Tesla A800 GPU with a random seed of 42. We set $\rho=1$ and initialize $\boldsymbol{\sigma}=0$ as a stress-free neutral state. We adopt SSIM~\cite{ssim}, PSNR~\cite{psnr}, and LPIPS~\cite{lpips} as evaluation metrics to comprehensively measure structural similarity, image fidelity, and perceptual quality. We streamline GaussianFlow~\cite{gao2024gaussianflow} CUDA/C++ differentiable rasterizer by removing redundant gradient and projection computations in backpropagation while retaining regularization support, boosting training efficiency and stability.

\subsection{Comparisons}

To address Question 1 in the Problem Formulation defined in Sec.~\ref{sec:Preliminary}.3, we compare PIDG with state-of-the-art methods on the custom physics-driven datasets and existing datasets, with the best results in \textbf{bold} and the second-best \underline{underlined}. Notably, previous physics-embedded dynamic reconstruction methods~\cite{xie2024physgaussian, chu2022physics,wang2024physics,zhong2025reconstruction,jiang2025phystwinphysicsinformedreconstructionsimulation} relied on strict boundary conditions and required RGB-D or multi-view inputs, making it difficult to generalize to monocular dynamic synthetic or real-world scenarios.  

\subsubsection{Results on PIDG physics-driven dataset.}
% As shown in Tab.~\ref{exp:dynamic_nvs_pidg}, our method achieves the best average quantitative results in the PIDG datasets. Moreover, our method demonstrates superior performance and generalization in diverse dynamic physical scenarios. Tab. 2 shows that our method achieves a balance between training efficiency while maintaining satisfactory reconstruction quality. Referring to visualizations in Fig.~\ref{exp_pidg}, further reveal that Grid4D~\cite{xugrid4d}, despite using the hash grid to achieve higher motion capacity, lacks explicit physical modeling, and GaussianPredict~\cite{zhao2024gaussianprediction} performs well on static regions but fails to capture fluid and elastic dynamics because it relies on motion distillation. In contrast, PIDG preserves physical motions effectively with physics-informed material fields and still achieves satisfactory static reconstruction under efficiency optimizations.

% \input{tables/time_pidg}
% Tab.~\ref{tab:supp-speed} further indicates that it strikes a good trade-off between training efficiency and reconstruction quality.

\begin{figure}[h]
\centering
\includegraphics[width=0.95\columnwidth]{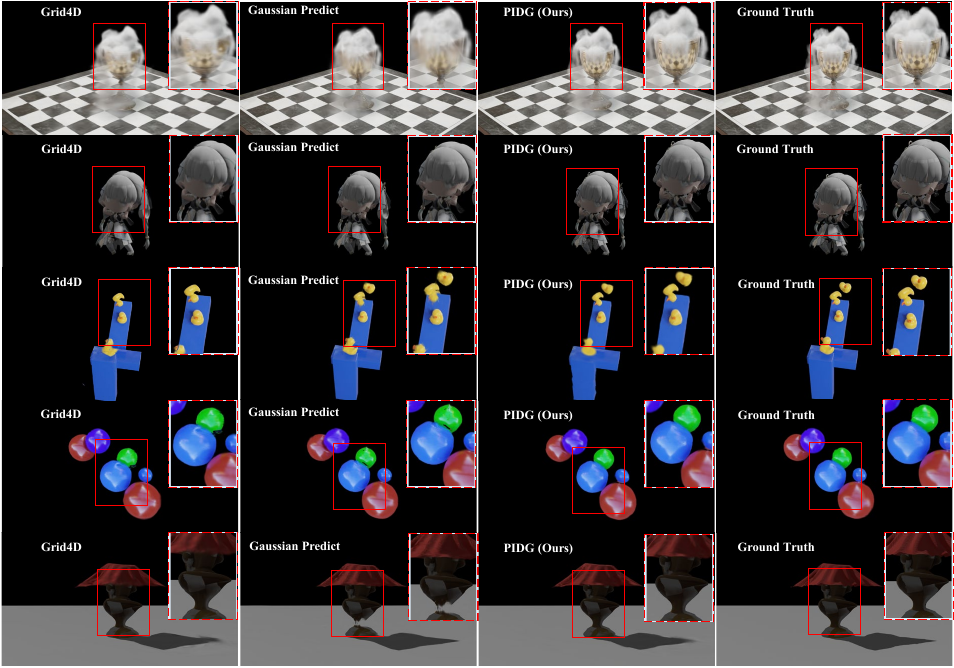} % Reduce the figure size so that it is slightly narrower than the column. Don't use precise values for figure width.This setup will avoid overfull boxes.
% \caption{The visualization results on the PIDG dataset, from left to right, Grid4D, GaussianPredict, PIDG (Ours), and the ground truth; from top to bottom, the scenes include fluid,  motion, soft-body, elastic collision, and cloth simulation.}
\caption{\textbf{The visual results on the PIDG dataset.}}
\label{exp_pidg}
\end{figure}

\begin{table}[ht]

\centering {\small
\begin{tabular}{@{}lcc@{}}
\toprule
Method (Monocular Inputs) & PSNR $\uparrow$ & MS-SSIM $\uparrow$ \\
\midrule
HyperNeRF~\cite{park2021hypernerf}                & 25.7 & 0.726 \\
4D-GS~\cite{wu20244d}                             & 26.9 & 0.798 \\
% D-3DGS~\cite{yang2024deformable}                 & 25.7 & 0.760 \\
MotionGS~\cite{zhumotiongs}                      & 25.2 & \underline{0.905} \\
GaussianPredict~\cite{zhao2024gaussianprediction} & 26.6 & 0.884 \\
SC-GS~\cite{huang2024sc}                         & 26.1 & 0.869 \\
Grid4D~\cite{xugrid4d}                           &  27.3 & 0.899  \\
MoDec-GS ~\cite{kwak2025modec} & 24.2 & 0.809 \\
D-2DGS ~\cite{D-2DGS} & 17.7 & 0.509\\

\midrule
PIDG ($w/o\ (\mathcal{L}_{\rm CMR}+\mathcal{L}_{\rm LPFM})$) & 27.5 & 0.901\\
PIDG ($w/o\ \mathcal{L}_{\rm LPFM}$) &   \underline{27.6}     & 0.902  \\
PIDG $\mathit{(Ours \ Full)}$   & \textbf{27.8} & \textbf{0.906}                             \\
\bottomrule
\end{tabular}
% \caption{\textbf{Average results on monocular scenes from the HyperNeRF real-world dataset}.  Full comparison results for each scenario are provided in Supp. Tab.~A4 and Fig.~A8.}
\caption{\textbf{Average results on the HyperNeRF dataset}. }
\label{tab:hypernerf_all}}
\end{table}

As shown in Fig.~\ref{exp_pidg}, our method models diverse dynamic motions while maintaining high-fidelity reconstruction in static regions, achieving coherent and generalizable motion representation. In contrast, Grid4D~\cite{xugrid4d}, although enhanced by hash grids, lacks physical modeling capability. GaussianPredict~\cite{zhao2024gaussianprediction}, based on motion distillation, performs well in static areas but fails to handle complex physical deformations. The mesh-dependent D-2DGS~\cite{D-2DGS} is limited to simple cases and struggles to generalize to complex synthetic or real-world scenes. As shown in Tab.~\ref{exp:dynamic_nvs_pidg}, our method achieves the best average results on the PIDG dataset and generalizes effectively across various dynamic scenarios. Specifically, PIDG ($w/o~(\mathcal{L}_{\rm CMR}+\mathcal{L}_{\rm LPFM})$) employs static-dynamic decoupling to guide deformable Gaussians, and PIDG ($w/o~\mathcal{L}_{\rm LPFM}$) uses a time-evolving material field constrained only by $\mathcal{L}_{\rm CMR}$, yielding consistent gains on most synthetic scenes except unconventional continuum cases. PIDG $\mathit{(Ours~Full)}$ further introduces Lagrangian particle flow matching to maintain coherent Gaussian flow across diverse scenes, significantly improving structural similarity in challenging fluid, soft-body, and elastic scenarios where deformation fields alone are insufficient. Additional results in the Supp. Sec.~D.1 show that our approach also achieves favorable training efficiency.

\subsubsection{Results on HyperNeRF real-world dataset.}

% As shown in Tab.~\ref{tab:hypernerf_all}, we achieve the best average quantitative results for monocular dynamic novel-view synthesis on the HyperNeRF real-world dataset~\cite{park2021hypernerf}. Notably, the incorporation of Lagrangian particle flow matching leads to a clear improvement in structural similarity, validating it as an effective supervision strategy without relying on any 3D prior. Although the quantitative gains brought by the introduction of the time-evolving material field are less pronounced in real-world scenarios than in synthetic datasets, we consider that pixel-level standard metrics may not fully capture the improvements in visual quality, which is consistent with the observation in~\cite{held2025triangle}. Our qualitative analysis in Fig.~\ref{exp_hyper} reveals significant improvements. In particular, the alignment of motion flow, velocity flow, and Gaussian flow is more coherent and intuitive, while MotionGS, also using optical flow supervision, has yielded relatively poorer visual results, especially in dynamic regions.

\begin{figure}[h]
\centering
\includegraphics[width=0.9\columnwidth]{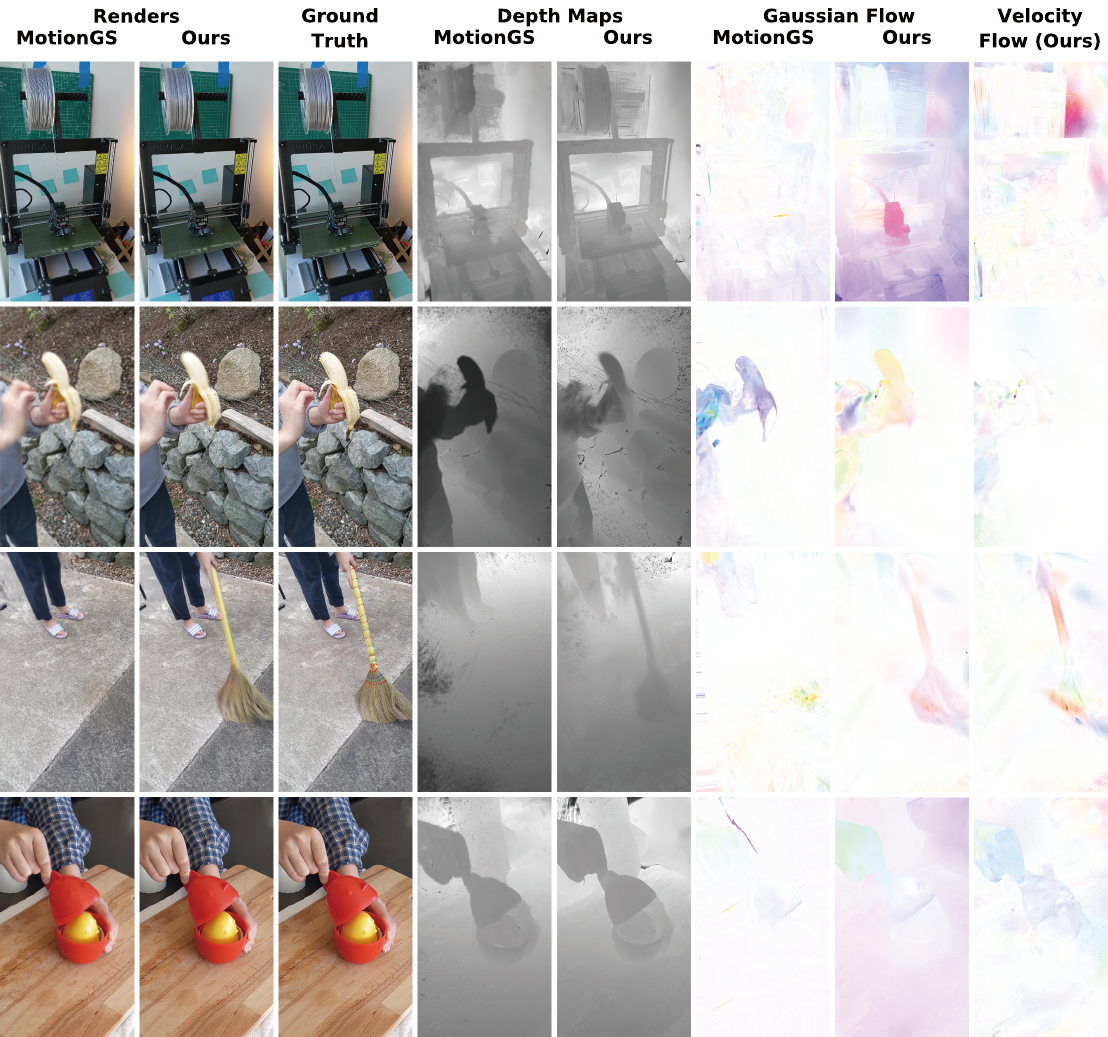} % Reduce the figure size so that it is slightly narrower than the column. Don't use precise values for figure width.This setup will avoid overfull boxes.
% \caption{\textbf{The visual comparison between PIDG (Ours) and MotionGS on the HyperNeRF dataset.} Our method achieves superior rendering quality, depth estimation, Gaussian flow. Moreover, and consistent velocity flows.}
\caption{\textbf{The visual comparison between PIDG (Ours) and MotionGS on the HyperNeRF dataset.}}
\label{exp_hyper}
\end{figure}

As shown in Tab.~\ref{tab:hypernerf_all}, we achieve the best average quantitative results for monocular dynamic novel-view synthesis on the HyperNeRF dataset. Notably, the incorporation of Lagrangian particle flow matching leads to a clear improvement in structural similarity, validating it as an effective supervision strategy without relying on ground-truth particle motion. Our qualitative analysis in Fig.~\ref{exp_hyper} further reveals significant improvements. In particular, the alignment of motion flow, velocity flow, and Gaussian flow is more coherent and intuitive, while MotionGS, also using optical flow supervision, has yielded relatively poorer visual results in dynamic regions.

% In contrast, MotionGS~\cite{zhumotiongs}, which also relies on optical flow supervision, fails to produce satisfactory results under the same experimental settings.

\subsection{Additional Ablation Study and Analysis.}

\begin{table}[t]
\centering 
\small{
\begin{tabular}{lcc}
% \midrule \multirow[b]{2}{*}{ Method } & \multicolumn{2}{c}{ Dynamic NVS}\\
% \cmidrule(lr){2-3}
% & PSNR \textbf{$\uparrow$} & SSIM \textbf{$\uparrow$}\\
% \midrule
\midrule
\multicolumn{1}{l}{Dynamic Novel-View Synthesis} & PSNR $\uparrow$ & SSIM $\uparrow$ \\
\midrule

% D-NeRF & 31.14 & 0.969  \\
% TiNeuVox & 32.74 & 0.972 \\
% Tensor4D & 27.44 & 0.942  \\
% K-Planes  & 31.41 & 0.970  \\
% 3D-GS & 23.40 & 0.930  \\
% 4D-GS & 36.92 & 0.984  \\
% D-3DGS  & 40.09 & 0.991  \\
GaussianPredict ~\cite{zhao2024gaussianprediction} & 40.58 & 0.992  \\
SC-GS ~\cite{huang2024sc} & 41.65 & 0.993 \\
Grid4D ~\cite{xugrid4d} & 42.00 & \underline{0.994} \\
MoDec-GS ~\cite{kwak2025modec} & 30.63 & 0.969 \\
D-2DGS ~\cite{D-2DGS} & 37.47 & 0.984 \\

GaussianPredict ($+ \ \mathcal{L}_{\rm CMR}$) & 40.93 & 0.993 \\
SC-GS ($+ \ \mathcal{L}_{\rm CMR}$) & 41.85 & \underline{0.994} \\
Grid4D ($+ \ \mathcal{L}_{\rm CMR}$) & \underline{42.10} & \underline{0.994} \\
PIDG ($w/o \ \mathcal{L}_{\rm LPFM}$) & \textbf{42.14} & \textbf{0.995}\\
% \midrule \multirow[b]{2}{*}{ Method } & \multicolumn{2}{c}{Future Prediction}\\
% \cmidrule(lr){2-3}
% & PSNR \textbf{$\uparrow$} & SSIM \textbf{$\uparrow$}\\
% \midrule

\midrule
\multicolumn{1}{l}{Future Prediction} & PSNR $\uparrow$ & SSIM $\uparrow$ \\
\midrule

% 4DGS & 24.67 & 0.929 \\
% D-3DGS & 23.84 & 0.926 \\
GaussianPredict - MLP & 24.64 & 0.932  \\
GaussianPredict - GCN & 25.22 & \underline{0.938} \\
GaussianPredict - MLP ($+ \ \mathcal{L}_{\rm CMR}$) & 24.96 & 0.936 \\
GaussianPredict - GCN ($+ \ \mathcal{L}_{\rm CMR}$)  & \textbf{26.04} & \textbf{0.942} \\
PIDG ($w/o\ (\mathcal{L}_{\rm CMR}+\mathcal{L}_{\rm LPFM})$) &  25.16 & 0.936 \\
  
PIDG ($w/o\ \mathcal{L}_{\rm LPFM}$) & \underline{25.36} & \underline{0.938} \\

\hline

\end{tabular}}
% \caption{\textbf{Average results on the D-NeRF synthetic dataset}, validating the effectiveness of introducing a time-evolving material field constrained solely by $\mathcal{L}_{\rm CMR}$ for dynamic novel-view synthesis and future prediction.}
\caption{\textbf{Average results on the D-NeRF synthetic dataset.}}
\label{exp:dnerf_all}
\end{table}

\begin{figure}[ht]
\centering
\includegraphics[width=1.0\columnwidth]{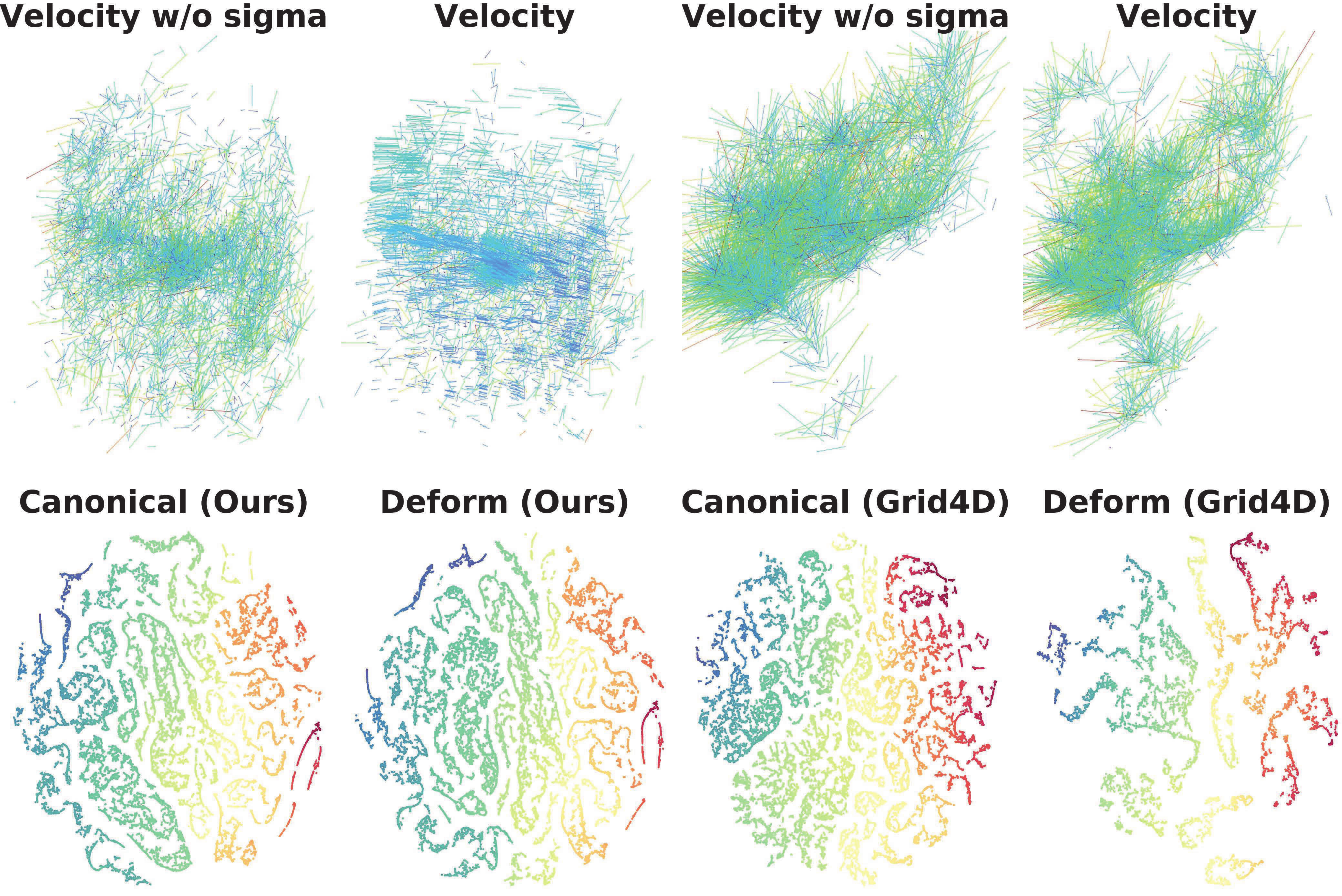} % Reduce the figure size so that it is slightly narrower than the column. Don't use precise values for figure width.This setup will avoid overfull boxes.
% \caption{\textbf{Analysis of consistency on PIDG dataset:} the first row analyzes the impact of the constitutive relation on material field velocities, and the second row shows the t-SNE of Gaussians in canonical and deformation spaces.}
\caption{\textbf{Constitutive relation and t-SNE analysis of consistency on PIDG dataset.}}
\label{vis_pidg}
\end{figure}

We study how velocity $\boldsymbol{v}$ and constitutive stress $\boldsymbol{\sigma}$ together ensure physical consistency through the Cauchy momentum residual $\mathcal{L}_{\rm CMR}$. As shown in Fig.~\ref{vis_pidg}, removing $\boldsymbol{\sigma}$ simplifies $\mathcal{L}_{\rm CMR}$ to the continuity constraint $\nabla\!\cdot\!\boldsymbol{v}=0$, which enforces smoothness but cannot describe real material interactions in fluids and elastic collisions. We compare t-SNE embeddings of Gaussian particles from PIDG and Grid4D in canonical and deformation spaces. Our method preserves richer dynamics in the canonical space and maintains them after deformation through dynamic decoupling and inheritable embeddings of indices, velocities, and stresses during densification. To test generalization, we apply the time-evolving material field supervised only by $\mathcal{L}_{\rm CMR}$ as a plug-and-play module to various methods on the D-NeRF dataset, which contains simple dynamic scenes for future prediction. Since discontinuous viewpoints produce unreliable optical flow, $\mathcal{L}_{\rm LPFM}$ cannot be used. In this case, $\mathcal{L}_{\rm CMR}$ acts as a physically consistent regularizer, ensuring globally coherent dynamics without flow-based supervision. As shown in Tab.~\ref{exp:dnerf_all}, it yields clear improvements in dynamic novel-view synthesis and enhances future prediction through deformation-field extrapolation and GCN-based GaussianPredict. By examining render results and flow alignments, and empirically validating our approach across diverse scenes and baseline methods, we address Question 2 in Sec.~\ref{sec:Preliminary}.3 and underscore the benefits of our time-evolving material field for physical modeling and consistency enforcement overall. These experiments show that PIDG exhibits strong generalization across both synthetic and real dynamic scenes.

\section{Conclusion}
\label{sec:Conclusion}
% In this paper, we propose a Physics-Informed Deformable Gaussian Splatting method that models 3D Gaussian particles from a Lagrangian perspective and introduces a time-evolving material field to unify constitutive motion relations. It provides effective supervision for spatiotemporal structure and intrinsic physical properties modeling. Our approach achieves strong physical consistency and generalization in monocular dynamic novel view synthesis. 

In this paper, we proposed Physics-Informed Deformable Gaussian Splatting that models 3D Gaussian particles from a Lagrangian perspective and introduces a time-evolving material field to unify the constitutive laws. This material field serves as a physics-based inductive supervision, facilitating the joint modeling of spatio-temporal structure and intrinsic physical properties. Experiments show that our approach achieves high physical consistency and strong generalization in monocular dynamic novel-view synthesis.

% Nevertheless, our method has certain limitations: the optimization-based reconstruction is relatively inefficient, and the PINN-based constitutive model primarily serves as a regularization constraint rather than fully capturing all physical properties. We speculate that future work could explore feed-forward dynamic reconstruction and physical simulation for real-time applications.

\section*{Acknowledgements}
This work was supported by the Fundamental and Interdisciplinary Disciplines Breakthrough Plan of the Ministry of Education of China, the National Natural Science Foundation of China (NSFC) grant 12425113, and the Natural Science Foundation of Jiangsu Province under Grant BK20240462. We also acknowledge support from the Key Laboratory of the Ministry of Education for Mathematical Foundations and Applications of Digital Technology, University of Science and Technology of China.

% This work was supported in part by the National Natural Science Foundation of China (NSFC) Major Research Plan on Interpretable and General Purpose Next-generation Artificial Intelligence under Grant No. 92370205, NSFC Grant 12425113, the Natural Science Foundation of Jiangsu Province under Grant BK20240462. We also acknowledge support from the Key Laboratory of the Ministry of Education for Mathematical Foundations and Applications of Digital Technology, USTC.
% \bigskip
% \noindent Thank you for reading these instructions carefully. We look forward to receiving your electronic files!

\appendix
\renewcommand{\thefigure}{A\arabic{figure}}
\renewcommand{\thetable}{A\arabic{table}}
\setcounter{figure}{0}
\setcounter{table}{0}

\section*{Supplementary Material}
\label{sec:Appendix}

\section{Evolution of Different Constitutive Laws}

As detailed in Sec.~4.2, our time-evolving material field embeds continuum mechanics principles into the Gaussian framework. Here we provide supporting derivations of the constitutive tensors for rigid bodies, isotropic elastic solids, and perfect fluids from a unified low-energy effective field theory (EFT) perspective~\cite{nicolis2015zoology,crossley2017effectivefieldtheorydissipative,Glorioso:2017fpd}. These constitutive laws define the Cauchy stress tensor \(\sigma_{ij}\), which enters the momentum balance equation
\begin{equation}
    \rho\left(\frac{\partial v_j}{\partial t}+v_i\partial_i v_j\right)=\partial_i\sigma_{ij}+f_j,
    \label{cmr_appendix}
\end{equation}
thus forming the physical backbone of our material field.

For a continuum medium in \(d=3\) spatial dimensions, three scalar Goldstone fields
\begin{equation}
    \Phi^I(x^\mu)\,, \; I=1,2,3,\;x^\mu=(t,\mathbf{x}),
\end{equation}
are introduced to map each physical point \(x^\mu\) to its Lagrangian label \(\Phi^I\). In equilibrium, \(\langle\Phi^I\rangle = x^I\). The internal symmetries of the medium determine its mechanical behavior: global translations and \(\mathrm{SO}(3)\) rotations characterize solids, volume-preserving diffeomorphisms \(\Phi^I \to \xi^I(\Phi)\) with \(\det(\partial\xi/\partial\Phi)=1\) distinguish fluids by eliminating static shear modulus, and global Euclidean motions \(\mathrm{ISO}(3)\) define rigid bodies as the zero-strain limit.

The invariant tensors
\begin{equation} B^{IJ}\equiv\partial_\mu\Phi^I\,\partial^\mu\Phi^J,\;\;b\equiv\sqrt{\det B^{IJ}}
\end{equation}
are constructed using a mostly-plus metric \(g_{\mu\nu}\). At leading order, the EFT action is given by
\begin{equation}
S=\int d^4x\,\mathcal{L},\quad
\mathcal{L}=
\begin{cases}
F(B^{IJ}) & \text{for solids},\\[4pt]
F(b)      & \text{for fluids}.
\end{cases}
\end{equation}
here, \(F\) is an analytic function whose Taylor coefficients encode the elastic moduli. The stress–energy tensor obtained from variation with respect to \(g_{\mu\nu}\) is
\begin{equation}
T^{\mu\nu}=-\frac{2}{\sqrt{-g}}\frac{\delta S}{\delta g_{\mu\nu}}
=-2\frac{\partial F}{\partial B^{IJ}}\partial^\mu\Phi^I\partial^\nu\Phi^J + g^{\mu\nu}F.
\label{eq:energy momentum}
\end{equation}

In the non-relativistic limit, the spatial components \(T_{ij}\) correspond to the Cauchy stress tensor \(\sigma_{ij}\) in Eq. (\ref{cmr_appendix}).

Expanding about equilibrium \(\Phi^I = x^I+\pi^I\), the Eulerian displacement \(\pi^I(t,\mathbf{x})\) and the infinitesimal strain tensor
\begin{equation}
    e_{ij}=\tfrac{1}{2}\left(\partial_i\pi_j+\partial_j\pi_i\right)+\cdots
\end{equation}
are introduced. A quadratic Taylor expansion of \(F(B^{IJ})\) yields
\begin{equation}
\mathcal{L}=-\rho_0\left[\tfrac{1}{2}\dot{\pi}^2
-\tfrac{1}{2}c_L^2(\partial_i\pi_i)^2
-\tfrac{1}{2}c_T^2(\partial_{[i}\pi_{j]})^2\right]+\cdots,
\end{equation}
where \(c_L^2=\tfrac{\lambda+2\mu}{\rho_0}\) and \(c_T^2=\tfrac{\mu}{\rho_0}\) are longitudinal and transverse phonon speeds, and \(\lambda,\mu\) are the Lamé parameters. The resulting constitutive law is
\begin{equation}
    \sigma_{ij}=\lambda\,e_{kk}\delta_{ij}+2\mu\,e_{ij}.
\end{equation}

For perfect fluids, volume-preserving diffeomorphism symmetry restricts \(F\) to depend solely on \(b\). The stress–energy tensor becomes
\begin{equation}
    T^{\mu\nu}=(\varepsilon+p)u^\mu u^\nu+p\,g^{\mu\nu},
\end{equation}
with \(p(b)=bF'(b)-F\) and \(\varepsilon(b)=F(b)\). In the non-relativistic limit \(u^\mu\to(1,\mathbf{v})\), the stress tensor reduces to
\begin{equation}
    \sigma_{ij}^{\rm(ideal)}=-p\delta_{ij}.
\end{equation}

Including first-order derivative corrections yields the viscous stress
\begin{equation}
    \sigma_{ij}^{\rm(visc)}=2\eta\,\dot{e}_{ij}+\zeta\,\dot{e}_{kk}\delta_{ij},
\end{equation}
where \(\eta,\zeta\) are the shear and bulk viscosities.

Rigid body kinematics is recovered by setting \(\Phi^I(t,\mathbf{x})=R^I{}_J(t)x^J+a^I(t)\), where \(R\in SO(3)\). Internal stresses act as Lagrange multipliers enforcing \(e_{ij}=0\). Substituting into Eq.~\eqref{eq:energy momentum} gives the equations of motion
\begin{equation}
    \frac{d}{dt}(Mv_i)=F_i,\quad
\frac{d}{dt}(I_{ab}\omega_b)=\tau_a,
\end{equation}
with mass \(M\) and inertia tensor \(I_{ab}\).

In all cases, the derived stress tensors \(\sigma_{ij}\) naturally enter the Cauchy momentum equation
\begin{equation}
    \rho\left(\partial_t\boldsymbol{v}+(\boldsymbol{v}\cdot\nabla)\boldsymbol{v}\right)=\nabla\cdot\boldsymbol{\sigma}+\mathbf{f},
\end{equation}
providing a unified physical basis for the time-evolving material field formulation. In conclusion, within a unified low-energy EFT framework, we show that as the constitutive stress tensor $\sigma_{ij}$ changes form, the Cauchy momentum equation continuously morphs into distinct dynamical regimes, rigid-body mechanics, isotropic linear elasticity, and ideal/viscous fluid dynamics, thereby providing a single physical backbone for our time-evolving material field.

\section{Calculation of Rendered Flow and Flow Warping Strategy}

\begin{figure}[htbp]
\centering
\includegraphics[width=1.0\columnwidth]{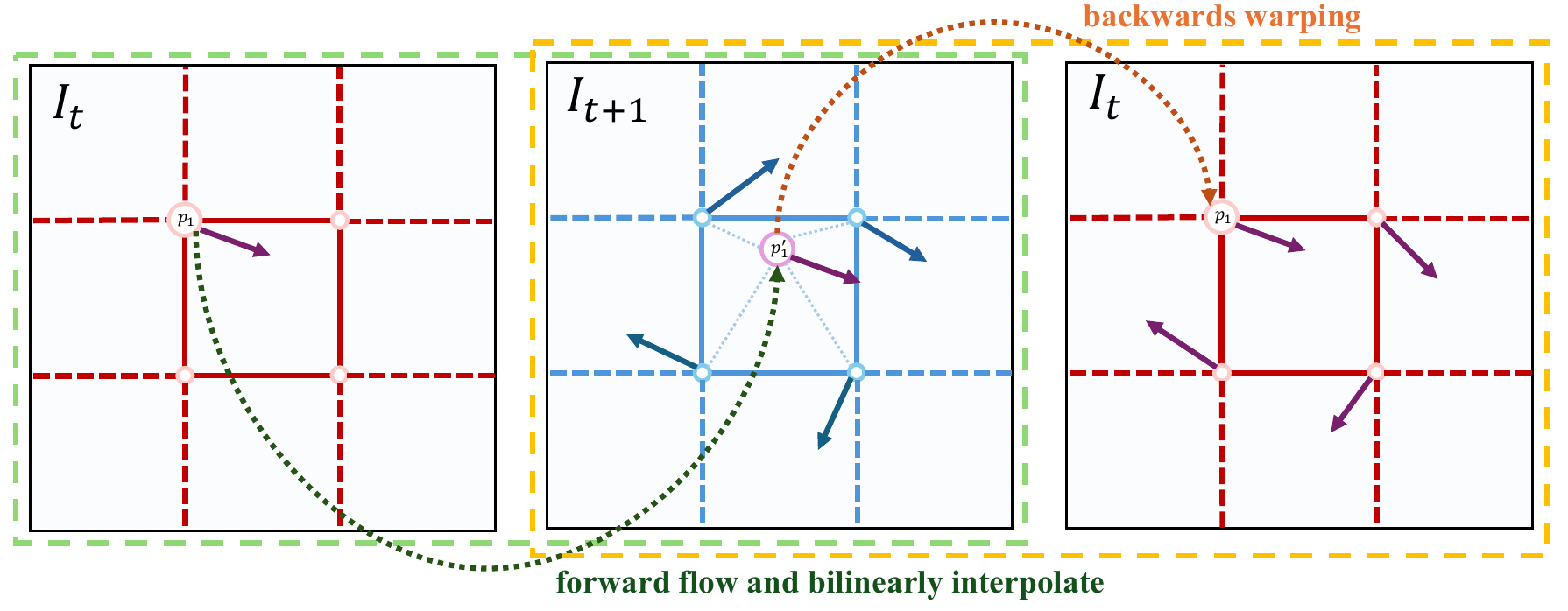}
\caption{\textbf{Backwards warping strategy}: unifying motion flow and Gaussian flow in a common view $I_t$.}
\label{warp}
\end{figure}

We adopt the formulation introduced in GaussianFlow~\cite{gao2024gaussianflow}, which explicitly propagates pixel positions through the full anisotropic support of each two-dimensional Gaussian primitive, rather than relying solely on the center position. Let a pixel at time $t$ be denoted as ${p}_1 \in \mathbb{R}^2$. The $i$-th Gaussian primitive is characterized by its mean ${\mu}_{i,t}$ and covariance matrix $\Sigma_{i,t}$ with singular value decomposition:

\begin{equation}
    \Sigma_{i,t} = U_i \Lambda_{i,t} V_i^{\top}, \quad \Lambda_{i,t} = \operatorname{diag}(\lambda_{i,t}^{(1)}, \lambda_{i,t}^{(2)}).
\end{equation}

Mapping ${p}_1$ to unit-circle coordinates yields
\begin{equation}
{p}_{\mathrm{std}, t} = \Lambda_{i,t}^{-\frac{1}{2}} ({p}_1 - {\mu}_{i,t}).
\end{equation}

GaussianFlow assumes that these normalized coordinates remain invariant across adjacent frames, that is, ${p}_{\mathrm{std}, t} = {p}_{\mathrm{std}, t+1}$. Given updated parameters $({\mu}_{i,t+1}, \Lambda_{i,t+1})$, the pixel position advected by the Gaussian flow becomes
\begin{align}
{p}_{\mathrm{std}, t+1} &= \Lambda_{i,t+1}^{-\frac{1}{2}} ({p}^{\mathrm{gaussian}}_{i,t+1} - {\mu}_{i,t+1}) \\
{p}^{\mathrm{gaussian}}_{i,t+1} &= {\mu}_{i,t+1} + \Lambda_{i,t+1}^{\frac{1}{2}} \Lambda_{i,t}^{-\frac{1}{2}} ({p}_1 - {\mu}_{i,t}).
\end{align}

The derivation of velocity flow follows the same procedure and is therefore omitted here for clarity.

Next, we discuss the detailed warping strategy involved in the computation of motion flow and Gaussian flow. Since object motion and camera motion are explicitly decoupled in our formulation, both the pseudo-ground-truth motion flow (derived from backward optical flow) and the Gaussian flow (computed from per-Gaussian transformations) must be compared in the same image plane. Accordingly, both the forward warping strategy employed in MotionGS~\cite{zhumotiongs} and our backward warping strategy require flow alignment; the only distinction lies in which field is warped. While MotionGS warps the Gaussian flow, our method warps the pseudo-ground-truth motion flow. In forward warping methods such as MotionGS, a motion flow is defined on the pixel of $I_t$ to record the three-dimensional displacement of scene points when viewed from camera $C_{t+1}$. A motion mask is then applied to the motion flow to restrict supervision to dynamic regions. However, the associated Gaussian flow is naturally computed in the coordinate frame of $C_{t+1}$. To align the Gaussian flow in $C_{t}$, MotionGS applies the camera flow to obtain ${p}_3 = {p}_1 + \operatorname{camera\_flow}({p}_1)$. The Gaussian flow is then bilinearly interpolated at ${p}_3$ and transferred back to ${p}_1$. We note that the forward alignment strategy has two drawbacks. First, camera flow encodes only camera motion and therefore only works for static objects, neglecting object motion in the 3D world, leading to systematic misalignment. Second, for pixels $p_3$ that are mapped outside the image domain, bilinear interpolation is ill-defined, producing stripe-like artifacts in the warped Gaussian flow. While masking Gaussian flow with the motion mask can hide these artifacts, it does not correct the underlying misregistration; the resulting erroneous gradients hinder convergence during optimization. 

In our framework, Gaussian flow is evaluated directly in the camera coordinate frame $C_t$, whereas the motion flow, derived from the backward optical flow, is defined in view $C_{t+1}$. To place both fields in a common domain, we warp the motion flow forward, as illustrated in Fig.~\ref{warp}. Concretely, for each pixel ${p}_1 \in I_t$, we compute its forward correspondence ${p}_1' = {p}_1 + \operatorname{forward\_flow}({p}_1)$ in $I_{t+1}$ and obtain the motion-flow vector at $p_1$ by bilinearly interpolating the motion flow at ${p}_1'$. Forward warping may introduce ghosting artifacts, as forward optical flow may misestimate the displacement of static regions. The interpolated value at location ${p}_4$ may thus contain residual motion originating from $C_{t+1}$. To suppress these artifacts. We employ a binary motion mask derived with SAMv2 \cite{ravi2024sam}, which confines the loss to pixels exhibiting significant motion, providing cleaner supervision and more stable convergence.

\begin{figure}[H]
\centering
\includegraphics[width=0.95\columnwidth]{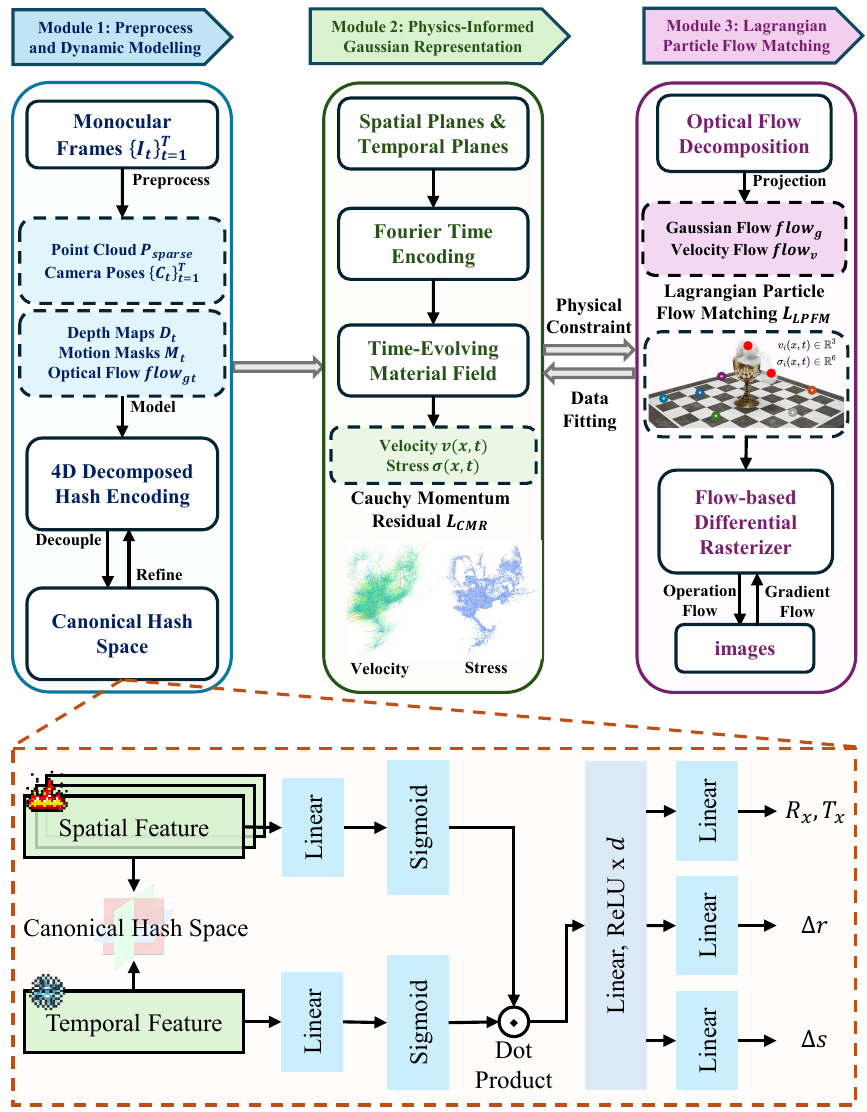} % Reduce the figure size so that it is slightly narrower than the column. Don't use precise values for figure width.This setup will avoid overfull boxes.
\caption{Top: Training pipeline of PIDG, including preprocessing and dynamic modeling, physics-informed Gaussian representation, and Lagrangian particle flow matching. Bottom: Network architecture of PIDG with 4D decomposed hash encoding in a decoupled canonical hash space.}
\label{warkflow}
\end{figure}

\section{Network Architecture of PIDG}

\subsection{Hash Encoding Configuration}
For spatial and temporal 3D hash grids, we set the maximum hash table size at \(2^{19}\). Each voxel across all grids encodes a feature vector of dimension 2. The spatial resolutions of the temporal grids are adapted to the scale of each scene: smaller scenes utilize higher resolutions to preserve fine details, while larger scenes utilize lower resolutions for efficiency. For the temporal dimension, the resolution is set between one-quarter and one-half of the total number of time samples in the dataset, balancing temporal fidelity with memory usage. To ensure consistent performance across datasets, all four hash grids are initialized with multi-resolution levels spanning from 16 to 2048. This configuration follows Grid4D~\cite{xugrid4d} by directly encoding 4D coordinates into overlapping 3D hash grids.

\subsection{Network Architectures}

Both the spatial MLP \(f_s\) and temporal MLP \(f_t\) adopted in Sec.~4.1 consist of a single fully connected layer followed by a ReLU activation. This shallow design suffices as the hash encoding already provides expressive features. As shown in Fig.~\ref{warkflow}, the multi-head directional attention decoder integrates three components: (i) a feature aggregator that combines outputs from \(f_s\) and \(f_t\) using a learned attention weight \(a\); (ii) a deformation predictor \(D(h)\), mapping the aggregated feature \(h\) to a translation \(T_x\), rotation increment \(\Delta r\), and scale update \(\Delta s\); (iii) The final deformation decoder has a depth of two, including the output layer, and features separate outputs for real-world and synthetic datasets. All hidden layers are 256 units wide and use ReLU activations. During training, we employ a two-stage procedure: Gaussian densification is jointly optimized with deformation initially, but is discontinued after 30,000 iterations to enable separate refinement of the static and dynamic parameters.

The time-evolving material field network \(f_\theta\) in Sec.~4.2 also adopts a multi-head MLP with depth 2 and width 256. The index embedding dimension \(H\) is set to 64. It jointly predicts the velocity \(\boldsymbol{v} \in \mathbb{R}^3\) and stress tensor \(\boldsymbol{\sigma} \in \mathbb{R}^6\) for each Gaussian particle. 

% The detailed training process is summarized in Algorithm~\ref{alg:pidg}.

\subsection{Training Settings and Optimization}
Our implementation builds on PyTorch 2.1 with CUDA 12.1. Our flow-based rendering backend, which is based on GaussianFlow~\cite{ gao2024gaussianflow}, extends 3D Gaussian Splatting with custom CUDA kernels for \(\nabla\cdot\boldsymbol{\sigma}\) and flow matching. Optimization employs Adam~\cite{kingma2017adammethodstochasticoptimization, hong2024ccspnet} with \(\beta=(0.9, 0.999)\). The background color is set to black during rendering. We use a hierarchical learning rate schedule similar to D-3DGS~\cite{yang2024deformable}, where the learning rate of the MLP decoder is adapted to the scale of each scene. In contrast, the grid hash encoders are trained with a learning rate 10-50 times larger than that of the MLP decoder to accelerate convergence in high-frequency regions. The base learning rate for the MLP is \(2\times10^{-3}\), exponentially decayed by 0.1 every $20k$ iterations. Following the experimental protocols of D-3DGS~\cite{yang2024deformable} and Grid4D~\cite{xugrid4d}, we select for each comparison method its best result within the same number of training iterations to ensure a fair comparison.

\subsection{Other Implementation Details}

\subsubsection{Details of Flow-based Differentiable Rasterizer.}
We optimize the GaussianFlow CUDA/C++ rasterizer for efficient backward flow and depth gradients. (1) In \texttt{computeColorFromSH()}, saturated SH channels (\texttt{clamped[3*idx+0..2]}) are zeroed to skip useless atomics. (2) In \texttt{computeCov2DCUDA()}, we merge \texttt{dL\_dconics} and \texttt{dL\_conic\_2D} into a single denom/adjugate pass. (3) Projection-coordinate gradients are written atomically to the 3D-mean buffer directly in \texttt{preprocessCUDA()} and \texttt{renderCUDA()}.

{\small
\begin{verbatim}
atomicAdd(&dL_proj_2D[global_id * 2 + 0], …);
atomicAdd(&dL_proj_2D[global_id * 2 + 1], …);
\end{verbatim}
}

This lets motion and depth gradients flow directly back to each Gaussian’s mean without a separate kernel. To prune unnecessary work at high primitive counts, \texttt{renderCUDA()} now skips projection and opacity writes whenever a Gaussian’s contribution is trivial, such as $\alpha<1/255$ or out-of-bounds. Finally, all kernels accumulate depth and optical flow regularization losses into the same gradient buffers. Overall, these CUDA-centric kernel fusions and atomic reductions collapse redundant backwards work and buffers. By exploiting warp-synchronous programming and shared-memory atomics, we achieve a leaner graph, lower peak memory, and faster steps than a pure framework-level autograd pipeline.

\subsubsection{Details of Custom Physics-Driven PIDG Dataset.}

The custom physics-driven PIDG dataset was constructed in Blender using its native physics solvers and rendering pipeline. Each scene was designed to represent a distinct physical phenomenon for evaluating dynamic novel-view synthesis. All scenes were rendered using the \textit{"Cycles"} engine at a resolution of 1920 × 1080, and resized to 1600 × 900 during training. Each scene contains 150 frames, corresponding to approximately 6 seconds of continuous monocular views at 25 FPS. The camera poses were uniformly sampled along a smooth orbital trajectory surrounding the scene, implemented in Blender by animating an empty object and parenting the camera to it with a \textit{"Track To"} constraint, ensuring smooth, centre-facing motion. Camera intrinsics and extrinsics were fully exported alongside the RGB frames to support dynamic reconstruction. Dynamic masks were generated by disabling the visibility of static objects during a secondary render pass in Blender. To ensure compatibility, all scenes were rendered and exported using the BlenderNeRF plugin~\cite{Raafat_BlenderNeRF_2024}, which provided the associated JSON files for camera poses. We adopted Blender’s default physics solver settings for each physical category across all scenes to ensure realistic yet reproducible physical behavior. Details of the scenes are as follows:

\begin{itemize}
\item \textbf{Elastic Mechanics (Balls Reaction):} Simulated with multiple rigid spheres in a bounded domain using \textit{"rigid body solver"}. Initial velocities were randomized to induce collision chains. The rigid body world was configured with the default mass = 1.0, bounciness = 0.6, and friction = 0.5. Materials were configured to emphasize contact points and rebound trajectories.

\item \textbf{Cloth Simulation (Mechanics Cloth):} Implemented via \textit{"Cloth Physics modifier"}. A grid mesh was pinned at selected vertices and subjected to gravity and wind forces. We used default cloth settings, including structural stiffness = 5.0, bending stiffness = 0.5, and air damping = 1.0, with self-collision enabled. The friction coefficient was set to 5.0 to ensure realistic folding and wrinkling as the cloth interacted with rigid bodies.

\item \textbf{Basic Exercises (Motion Kuro):} The models, including open-source human character models, were directly bound to predefined motion paths, such as sinusoidal translations and circular rotations, using a \textit{keyframe animation system}. No additional physical parameters were configured in this scene, as the motion was purely kinematic and did not rely on rigid body simulation.

\item \textbf{Soft Body (Rubber Duck):} Generated using \textit{"Soft-Body physics system"}. The duck model, adapted from open-source assets with physically plausible materials, was dropped from a height onto a rigid surface to produce elastic deformations. The soft body solver used default parameters: goal stiffness = 0.7, goal damping = 0.2, and bending stiffness = 4.0. Collision margin was set to 0.015 m, and edge spring stiffness to 0.5 to approximate a soft, resilient response.

\item \textbf{Fluid Smoke (Dry Ice):} Simulated using \textit{"Mantaflow fluid solver"} in smoke mode. A volumetric emitter was placed in a confined domain with default temperature = 1.0, vorticity = 2.0, and turbulence strength = 0.75. The emitter produced dense smoke that interacted with obstacles and evolved complexly over time. 
\end{itemize}

\begin{figure}[h]
\centering
\includegraphics[width=0.95\columnwidth]{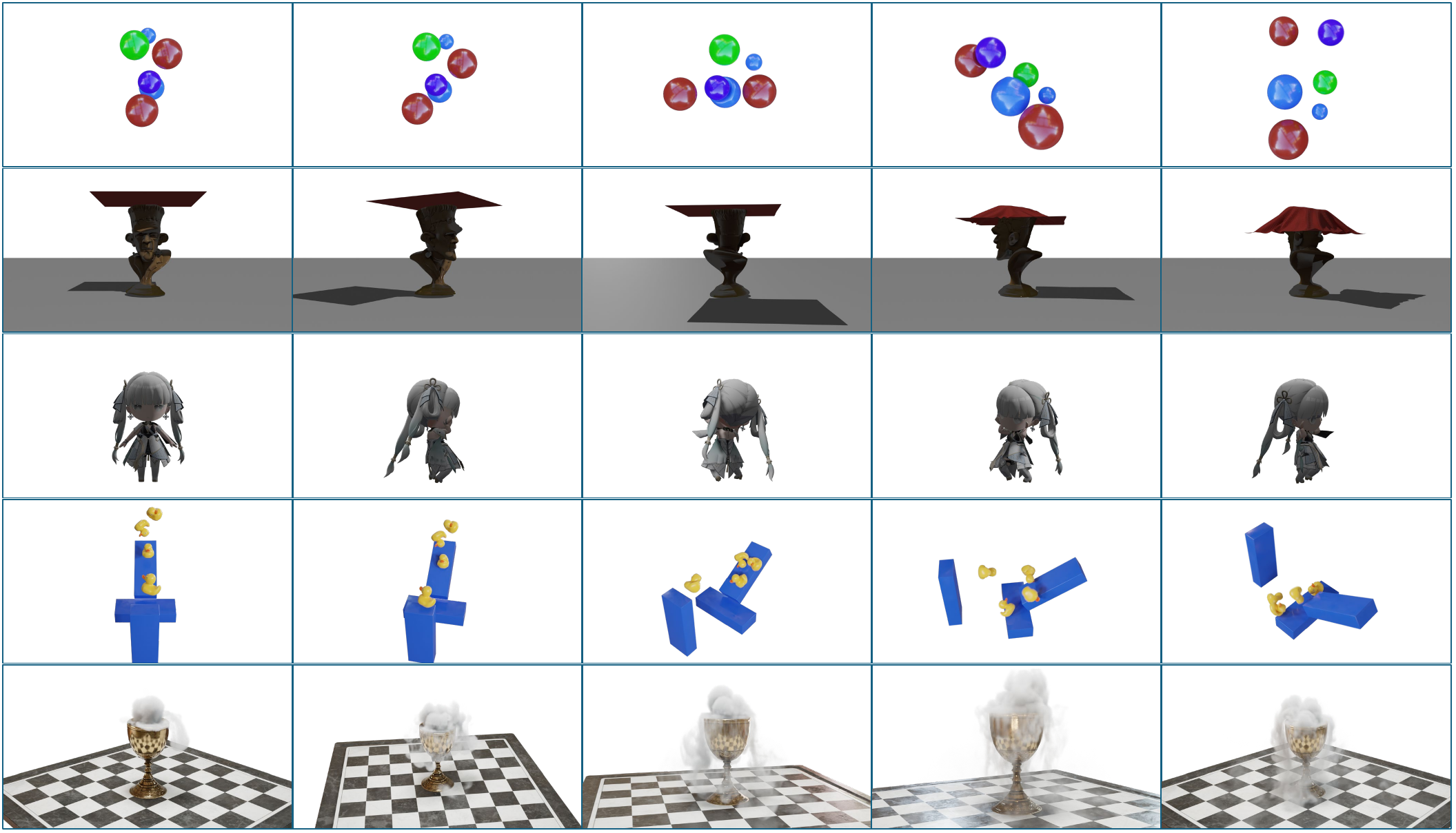} % Reduce the figure size so that it is slightly narrower than the column. Don't use precise values for figure width.This setup will avoid overfull boxes.
\caption{\textbf{Sample frames from our physics-driven  PIDG synthetic dataset.} From top to bottom: elastic mechanics (balls reaction), cloth simulation (mechanics cloth), basic exercises (motion kuro), soft-body reaction (rubber duck), and smoke simulation (dry ice). These dynamic scenes were created in Blender using a physics solver to address the lack of various physical scenarios in existing datasets. Each scene comprises approximately 150 frames.}
\label{pidg_datasets}
\end{figure}

A schematic of the sequential sampling of the experimental dataset is shown in Fig.~\ref{pidg_datasets}. Additionally, we include sample continuous video frames in Fig.~\ref{pidg_datasets_continuous} at the end of the supplementary materials to help reviewers better understand the dataset structure.

\subsubsection{Block-Sampled Cauchy Momentum Residual.}

To preserve consistent velocity and stress embeddings across subdivisions, we keep each child particle’s index identical to its parent’s. This avoids any post-subdivision identity recovery via nearest-neighbor search or clustering. However, in large-scale scenes, which may contain hundreds of thousands to millions of particles, computing the full Cauchy momentum residual, including stress-divergence terms for every particle in one backwards pass, can exhaust GPU memory due to the retention of a massive computation graph. To address this, we implement a two-stage process: $(a)$ Divide particles into blocks, computing each block’s Cauchy momentum residual and freeing its graph before aggregating weighted losses. $(b)$ Optionally, sample a small fixed-index subset of particles each iteration for the physical loss, while retaining full forward passes for up-to-date velocity and stress. This block-sampled Cauchy momentum residual strategy significantly reduces peak memory usage while maintaining accurate physics-informed constraints.

% \begin{itemize}
%   \item \textbf{Block-wise computation.} We partition the particle set into smaller blocks, compute each block’s residual separately,  release its intermediate graph, and then aggregate the block losses with block-size weighting.
%   \item \textbf{Optional sub-sampling.} For further memory savings, we can sample a small random subset of particles each iteration, compute the physical loss only on this subset, and add it to the total loss. Since these sampled residuals do not alter the network or particle indices, we can still perform a full forward pass over all particles at any time to retrieve up-to-date velocity and stress.
% \end{itemize}

% In our implementation, child particles reuse their parent indices and embeddings, preserving identity, velocity, and stress. To avoid GPU memory exhaustion from building a backwards graph over millions of particles for the full Cauchy momentum loss, we either partition particles into blocks, computing and freeing each block’s loss immediately, or sample a subset for loss evaluation. These strategies leave the network and indices unchanged, while a full forward pass over all particles still retrieves their velocity and stress fields.

\subsubsection{Scaling Threshold Adjustment Strategy.}

\begin{figure}[htbp]
\centering
\includegraphics[width=1.0\columnwidth]{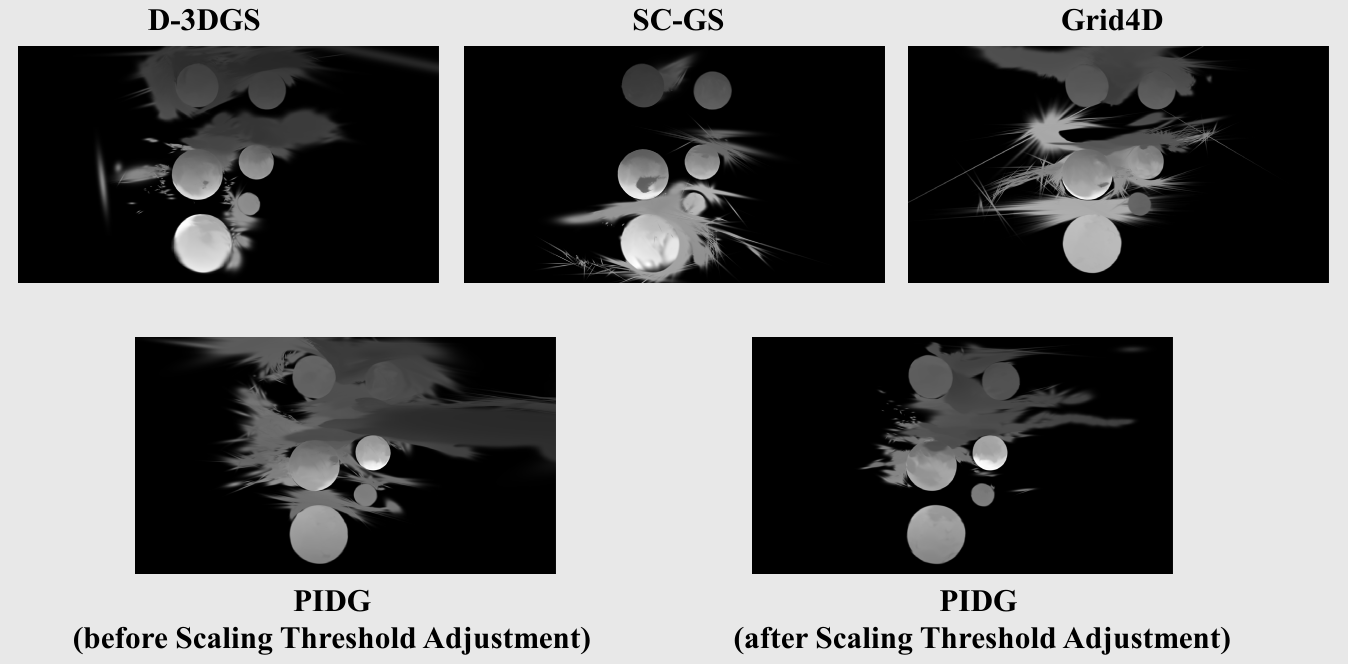} % Reduce the figure size so that it is slightly narrower than the column. Don't use precise values for figure width.This setup will avoid overfull boxes.
\caption{\textbf{Scaling threshold adjustment strategy.}}
\label{ablation_depth}
\end{figure}

Training on the PIDG dataset reveals feathery artifacts in optical flow and depth maps. These arise from high-opacity Gaussians whose colors are fitted to near-black background pixels, which are not sufficiently constrained by the photometric loss alone. Consequently, they destabilize velocity optimization, corrupt flow alignment, and exacerbate depth-prediction errors. Due to the lack of geometric cues from the single-camera trajectory, depth predictions are generally poor. Although D-NeRF achieves better depth using random 360-degree views, this approach is impractical for most real-world captures. While these noisy Gaussians scarcely degrade rendered appearance, they impair depth estimation and flow alignment. To mitigate this, we introduce a scaling threshold adjustment strategy. The original 3DGS method sets the scale threshold based on the radius of the camera path's bounding sphere, which is overestimated in PIDG. We reduce the threshold from 0.1 to 0.015 to better filter out large, noisy Gaussians. As shown in Fig.~\ref{ablation_depth}, this reduces depth noise and improves geometric reconstruction, confirming the effectiveness of the proposed adjustment.

\section{Additional Comparisons}
\begin{table*}[t]
\label{time}
% \caption{Average training efficiency on the PIDG dataset.}
  \centering
  \small
  \setlength{\tabcolsep}{3mm}
  \begin{tabular}{lcccccc}
    \toprule
    Model      & D-3DGS & SC-GS & Grid4D & GaussianPredict & D-2DGS & PIDG (Ours) \\
    \midrule
    Time & 80 $min$ & 75 $min$   & \textbf{67} $min$    & 101 $min$ & 85 $min$ & \underline{72} $min$  \\
    Memory & 7.2 $GB$  & \textbf{4.3} $GB$   & \underline{6.0} $GB$    & 6.7 $GB$ & 18.2 $GB$ & 6.2 $GB$  \\
    Nums & $\sim$ 110 $ k$  & $\sim$ 95 $ k$   & $\sim$ 100 $ k$    &  $\sim$ 120 $ k$  $ $ & $\sim$ 150 $ k$ & $\sim$ 85 $ k$  \\
    FPS & $\sim$ 180 & $\sim$ 210 & $\sim$ \underline{240} & $\sim$ 110 & $\sim$ 210   & $\sim$ \textbf{250}  \\
    \bottomrule
  \end{tabular}
\caption{\textbf{Average training efficiency on the PIDG dataset with a single 80GB NVIDIA Tesla A800 GPU.} We evaluated the training time, memory consumption, maximum point‑cloud size, and FPS of multiple models.}
\label{tab:supp-speed}
\end{table*}

\begin{table*}[!bh]
\centering
\small
\setlength{\tabcolsep}{1.1mm}{
\begin{tabular}{lcccccc}
\toprule
Model & Metric & $L=8, d=0$ & $L=16, d=0$ & $L=32, d=0$ & $L=32, d=1$ & $L=32, d=2$ \\
\midrule
\multirow{2}{*}{Grid4D} 
& PSNR~$\uparrow$ & 29.43 & 30.25 & \underline{30.32} & 30.14 & 30.07 \\
& SSIM~$\uparrow$ & 0.946 & \underline{0.956} & \underline{0.956} & 0.950 & 0.948 \\
\midrule
\multirow{2}{*}{PIDG (Ours)}
& PSNR~$\uparrow$ & 30.04 & 30.71 & \textbf{30.95} & 30.88 & 30.15 \\
& SSIM~$\uparrow$ & 0.955 & 0.961 & \textbf{0.965} & 0.958 & 0.954 \\
\bottomrule
\end{tabular}}
\caption{\textbf{Additional ablation average results of model architecture on the PIDG dataset.} Following and comparing the Grid4D~\cite{xugrid4d}, we change the depth $d$ of the multi-head decoder, as well as the max level number $L$ of the temporal grid hash encoder. The best settings for PIDG are in \textbf{bold} and the best settings for Grid4D are \underline{underlined}.}
\label{tab:supp-ablation-d-nerf}
\end{table*}

\begin{figure*}[htbp]
\centering
\includegraphics[width=1.8\columnwidth]{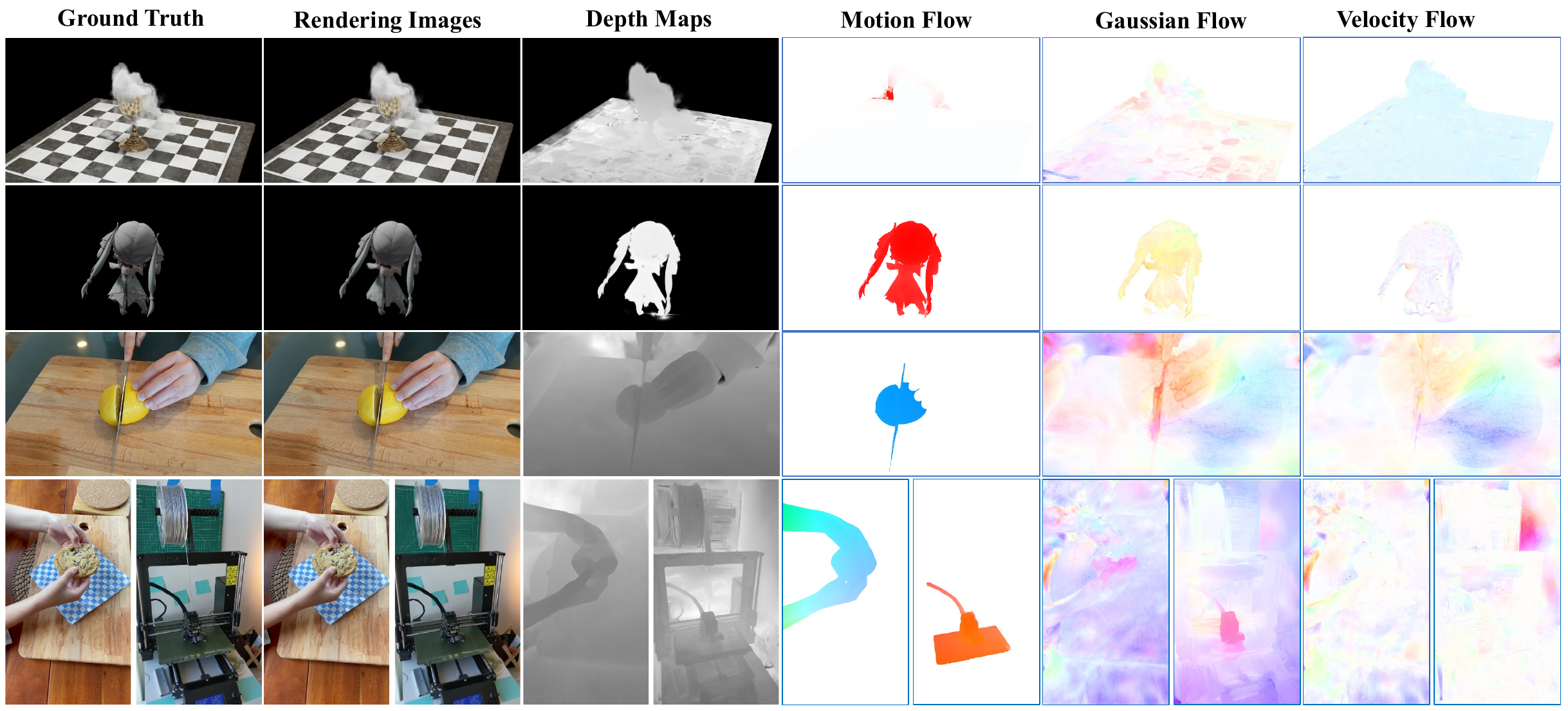} % Reduce the figure size so that it is slightly narrower than the column. Don't use precise values for figure width.This setup will avoid overfull boxes.
\caption{\textbf{The visualization of Physics-Informed Deformable Gaussian Splatting (PIDG) method}. Our approach delivers compelling rendering performance in monocular dynamic novel view synthesis while additionally supporting depth prediction. By leveraging motion flow and velocity flow, it further ensures enhanced consistency in the aligned Gaussian flows. We showcase representative scenes from the PIDG physics-driven synthetic dataset and the HyperNeRF real-world dataset~\cite{park2021hypernerf} to demonstrate the effectiveness of our method.}
\label{pidg_hyper_all}
\end{figure*}

We further extend the experiments beyond the main text to more rigorously validate the improved generalization and consistency of our method on monocular dynamic novel view synthesis. As shown in Fig.~\ref{pidg_hyper_all}, we not only produce high-quality dynamic novel view renderings, but also predict depth maps, incorporating motion flow and velocity flow, the aligned Gaussian flow exhibits visually consistent behavior.

\subsection{Additional results on custom PIDG datasets.}

\begin{figure*}[htbp]
\centering
\includegraphics[width=1.8\columnwidth]{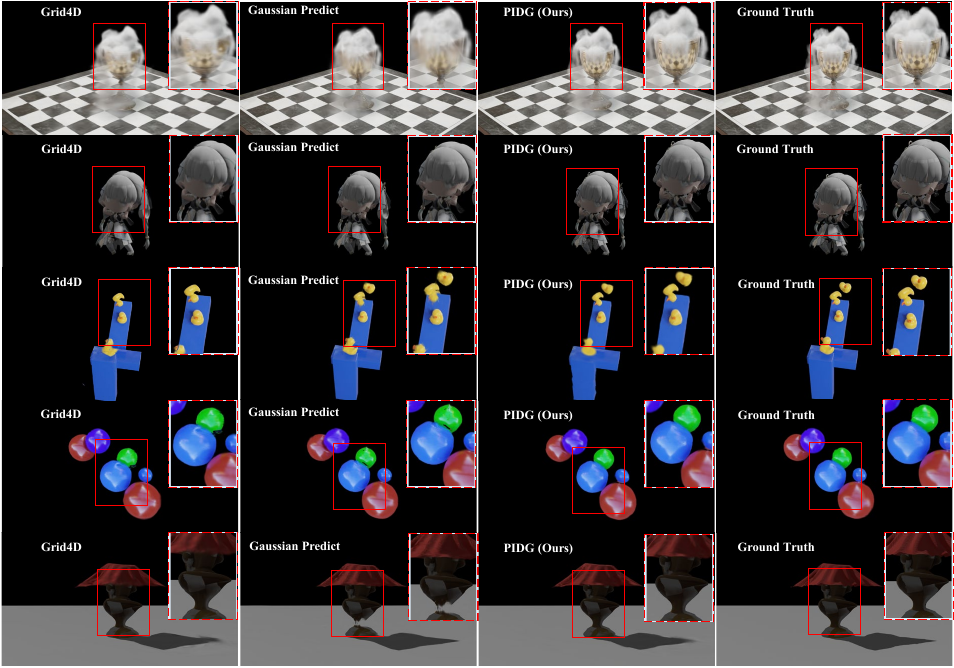} % Reduce the figure size so that it is slightly narrower than the column. Don't use precise values for figure width.This setup will avoid overfull boxes.
\caption{\textbf{The comparison of visual results on the PIDG dataset (higher-resolution version)}, from left to right, Grid4D, GaussianPredict, PIDG (Ours), and the ground truth; from top to bottom, the scenes include fluid simulation,  basic exercises, soft-body reaction, elastic mechanics, and cloth simulation.}
\label{exp_pidg_high}
\end{figure*}

We evaluate our method on the custom physics-driven PIDG dataset, which covers diverse dynamic scenes including fluid, soft-body, elastic, and cloth simulations, to validate the reconstruction quality and generalization. As shown in Fig.~\ref{exp_pidg_high}, our method not only achieves superior reconstruction quality and consistency on the dynamic regions, but also maintains high visual fidelity in the static areas, producing visually coherent and physically plausible results that outperform prior methods in both aspects.

To bridge these observations, we note that the static-dynamic decoupling and the optimized gradient flow for optical motion estimation work synergistically: the former reduces redundant computation by separating static and dynamic regions, thereby alleviating the burden of deformation field estimation, while the latter suppresses gradients on unnecessary Gaussian particles through strong flow-based supervision, cutting memory overhead. Together, they achieve the efficiency-quality trade-off reported in Tab.~\ref{tab:supp-speed}. This synergy allows the model to scale to varied dynamic scenes without sacrificing reconstruction fidelity. We further vary the multi-head decoder depth \(d\) and the maximum level \(L\) of the temporal grid hash encoder. As shown in Tab.~\ref{tab:supp-ablation-d-nerf}, deeper models can degrade, which we attribute to training difficulties of deep MLPs, and we find \(L=32, d=0\) performs best. Combined with the analyses in the main text, we conclude that PIDG achieves superior reconstruction quality and generalization in monocular dynamic novel view synthesis, particularly across diverse physical scenarios.

\subsection{Detailed results on HyperNeRF datasets.}

\begin{table*}[ht]
\centering
% \fontsize{6.5}{8}\selectfont
% \vspace{-0.5em}
\resizebox{\linewidth}{!}{
\begin{tabular}{@{}l *{5}{cc} c@{}}
\toprule
  & \multicolumn{8}{c}{\textbf{Vrig Datasets (Only Use Single View)}} \\    % 第一层分组行
\cmidrule(lr){2-9}
\multirow[b]{2}{*}{Method}
  & \multicolumn{2}{c}{3D Printer}
  & \multicolumn{2}{c}{Chicken}
  & \multicolumn{2}{c}{Broom}
  & \multicolumn{2}{c}{Peel Banana} \\
\cmidrule(lr){2-3}\cmidrule(lr){4-5}\cmidrule(lr){6-7}\cmidrule(lr){8-9}
  & PSNR $\uparrow$ & MS-SSIM $\uparrow$
  & PSNR $\uparrow$ & MS-SSIM $\uparrow$
  & PSNR $\uparrow$ & MS-SSIM $\uparrow$
  & PSNR $\uparrow$ & MS-SSIM $\uparrow$ \\
\midrule
HyperNeRF~\cite{park2021hypernerf}                           & 20.0 & 0.635 & 27.4 & 0.632 & 19.5 & 0.214 & 22.1 & 0.719 \\
% TiNeuVox~\cite{fang2022fast}                           & 22.8 & 0.73 & 28.2 & 0.79 & 21.3 & 0.31 & 24.4 & 0.64 \\
4D-GS~\cite{wu20244d}                              & \textbf{22.0} & 0.712 & 28.5 & 0.805 & 22.0 & 0.374 & \textbf{27.3} & 0.853 \\
D-3DGS ~\cite{yang2024deformable}                            & 20.5 & 0.643 & 22.8 & 0.614 & 20.5 & 0.348 & 26.0 & 0.832 \\
MotionGS ~\cite{zhumotiongs}                          & 21.0 & \textbf{0.817} & 25.5 & 0.919 & 21.2 & \textbf{0.777} & 24.2 & \textbf{0.936} \\
GaussianPredict ~\cite{zhao2024gaussianprediction}                   & 21.4 & 0.796 & 27.1 & 0.924 & 20.2 & 0.673 & 25.4 & 0.909 \\
SC-GS ~\cite{huang2024sc} & 19.1 & 0.764 & 22.8 & 0.783 & 21.8 & 0.711 & 26.0 & 0.918 \\
Grid4D  ~\cite{xugrid4d}                           & 21.5 & 0.799 &  28.5  &  0.940  & 21.8 & 0.741 &  26.0  &  0.910  \\
MoDec-GS ~\cite{kwak2025modec} & 21.1 & 0.773 & 26.5 & 0.894 & 21.3 & 0.663 & \underline{27.0} & 0.924 \\
D-2DGS ~\cite{D-2DGS} & 16.8 & 0.516 & 16.8 & 0.473 & 19.7 & 0.507 & 16.1 & 0.309\\
\midrule
PIDG ($w/o \ (\mathcal{L}_{\rm{CMR}} + \mathcal{L}_{\rm{LPFM}})$) & 21.6 & 0.801 & \underline{28.6}  & 0.943 & 21.9 & 0.741 & 26.3 & 0.914\\
PIDG ($w/o \ \mathcal{L}_{\rm{LPFM}}$) & 21.6 &  0.801 & \textbf{28.7}  & \underline{0.944} & \underline{22.1} & \underline{0.743} & 26.2 & 0.912    \\
PIDG $\mathit{(Ours \ Full)}$    & \underline{21.8} & \underline{0.810} & \textbf{28.7} & \textbf{0.946} & \textbf{22.2}  &  0.741  & \underline{27.0}  &  \underline{0.925}  \\
\midrule
  & \multicolumn{6}{c}{\textbf{Monocular Video Datasets}} & & \\    % ← 只跨6列
\cmidrule(lr){2-7}
\multirow[b]{2}{*}{Method}
  & \multicolumn{2}{c}{Cut Lemon}
  & \multicolumn{2}{c}{Split Cookie}
  & \multicolumn{2}{c}{Chocolate}
  & \multicolumn{2}{c}{Average} \\
\cmidrule(lr){2-3}\cmidrule(lr){4-5}\cmidrule(lr){6-7}\cmidrule(lr){8-9}
  & PSNR $\uparrow$ & MS-SSIM $\uparrow$
  & PSNR $\uparrow$ & MS-SSIM $\uparrow$
  & PSNR $\uparrow$ & MS-SSIM $\uparrow$
  & PSNR $\uparrow$ & MS-SSIM $\uparrow$ \\
\midrule
HyperNeRF  ~\cite{park2021hypernerf}                         & 31.8 & 0.958 & 30.9 & 0.968 & 28.0 & 0.959 & 25.7 & 0.726\\
% TiNeuVox   ~\cite{fang2022fast}                        & 28.6 & 0.96 & 28.9 & 0.97 & 26.8 & 0.95 &   &    \\
4D-GS   ~\cite{wu20244d}                           & 30.0 & 0.934 & 32.5 & 0.976 & 26.1 & 0.934 & 26.9 & 0.798    \\
D-3DGS ~\cite{yang2024deformable}                            & 30.0 & 0.952 & 32.7 & 0.975 & 27.5  &  0.956  & 25.7 & 0.760    \\
MotionGS ~\cite{zhumotiongs}                          &  28.0  &  0.939  &  29.4  &  0.979  &  26.8 &  \textbf{0.968}  & 25.2 & \underline{0.905}    \\
GaussianPredict ~\cite{zhao2024gaussianprediction}                   & 30.9 & 0.950 & 33.3 & 0.976 & 27.9 & 0.962 & 26.6 & 0.884    \\
SC-GS ~\cite{huang2024sc} & 31.4 & 0.965 & \textbf{33.5} & 0.978 & \textbf{28.4} & \underline{0.966} & 26.1 & 0.869 \\
Grid4D ~\cite{xugrid4d}                            & 31.8 & 0.960 &  33.2  &  0.980  & \underline{28.2} & 0.964 & 27.3 & 0.899\\
MoDec-GS ~\cite{kwak2025modec} & 29.3 & 0.937 & 17.7 & 0.520 & 26.7 & 0.953 & 24.2 & 0.809\\
D-2DGS ~\cite{D-2DGS}  & 21.1 & 0.728 & 14.4 & 0.316 & 18.7 &  0.717 & 17.7 & 0.509 \\
\midrule
PIDG ($w/o \ (\mathcal{L}_{\rm{CMR}} + \mathcal{L}_{\rm{LPFM}})$) & 32.5 & 0.966 & 33.3  & 0.979 & 28.1 & 0.964 & 27.5 & 0.901 \\
PIDG ($w/o \ \mathcal{L}_{\rm{LPFM}}$) & \underline{32.8} &  \underline{0.968} & \underline{33.4}  & \underline{0.981} & \underline{28.2} &  0.963 & \underline{27.6} & 0.902 \\
PIDG $\mathit{(Ours \ Full)}$  & \textbf{32.9} & \textbf{0.970} & \underline{33.4} & \textbf{0.983} &  \underline{28.2} & 0.965 &  \textbf{27.8} & \textbf{0.906}   \\
\bottomrule
\end{tabular}
}
\caption{\textbf{Quantitative dynamic novel view synthesis results on the HyperNeRF real-world dataset}. PIDG ($w/o\ (\mathcal{L}_{\rm CMR} + \mathcal{L}_{\rm LPFM})$) denotes using only the static-dynamic separated 4D decomposed hash encoding network; PIDG ($w/o\ \mathcal{L}_{\rm LPFM}$) denotes results without Lagrangian particle flow matching (only the time-evolving material field predicting particle velocity); PIDG $\mathit{(Ours\ Full)}$ denotes the full model. The best results are in \textbf{bold} and the second-best are \underline{underlined}.}
\label{exp:dynamic_nvs_hypernerf}
\end{table*}

We evaluate on the vrig subsets of HyperNeRF real-world datasets~\cite{park2021hypernerf}, including 3D Printer, Chicken, Broom, and Peel-Banana, using the left-view sequence as the monocular input. We further include highly deformable scenes, including Cut-Lemon, Split-Cookie, and Torchochocolate, for deeper analysis.

\begin{figure}[!htbp]
\centering
\includegraphics[width=1.0\columnwidth]{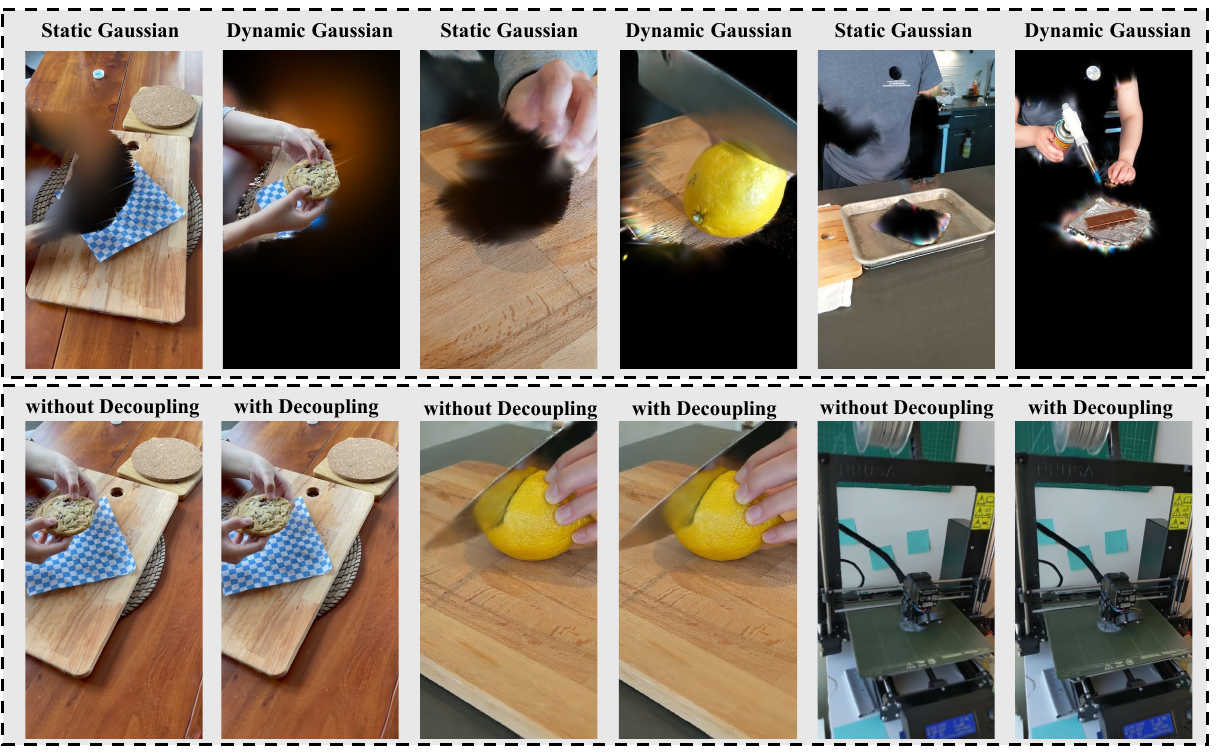} % Reduce the figure size so that it is slightly narrower than the column. Don't use precise values for figure width.This setup will avoid overfull boxes.
\caption{\textbf{The Effect of Applying Dynamic-Static Decoupling.} In the first row, we present the separate rendering results of the dynamic and static Gaussian splatting components after training. It can be observed that our method has reasonably achieved the separation of dynamic and static elements. In the second row, we demonstrate the impact of ablating the dynamic-static decoupling approach on rendering quality. Although standard metrics do not show significant improvement, the visual results exhibit higher clarity.}
\label{hyper_static}
\end{figure}

As shown in Table~\ref{exp:dynamic_nvs_hypernerf}, PIDG  ($w/o~\mathcal{L}_{\rm CMR}+\mathcal{L}_{\rm LPFM}$), which models Gaussian particle deformation using only 4D decomposed hash encoding with static-dynamic decoupling, shows varying performance across different scenes. This occurs mainly because subtle or unclear motions result in inaccurate motion masks, leading to less effective static-dynamic separation. As illustrated in Figure~\ref{hyper_static}, while these cases may not show significant improvements in quantitative metrics, we consistently observe better reconstruction quality in static regions after applying the decoupling approach. For scenes with distinct motion boundaries such as Cut-Lemon, the benefits are more obvious, with both visibly sharper details and measurable gains in evaluation scores. PIDG ($w/o~\mathcal{L}_{\rm LPFM}$), which introduces a time-evolving material field supervised solely by $\mathcal{L}_{\rm CMR}$, yields consistent yet moderate improvements; however, real footage exhibits more irregular motion than synthetic data, and scenes such as 3D Printer and Peel-Banana, whose dynamics are not well described by a continuum-mechanics framework, do not receive strong consistency constraints.

\begin{figure}[htbp]
\centering
\includegraphics[width=0.9\columnwidth]{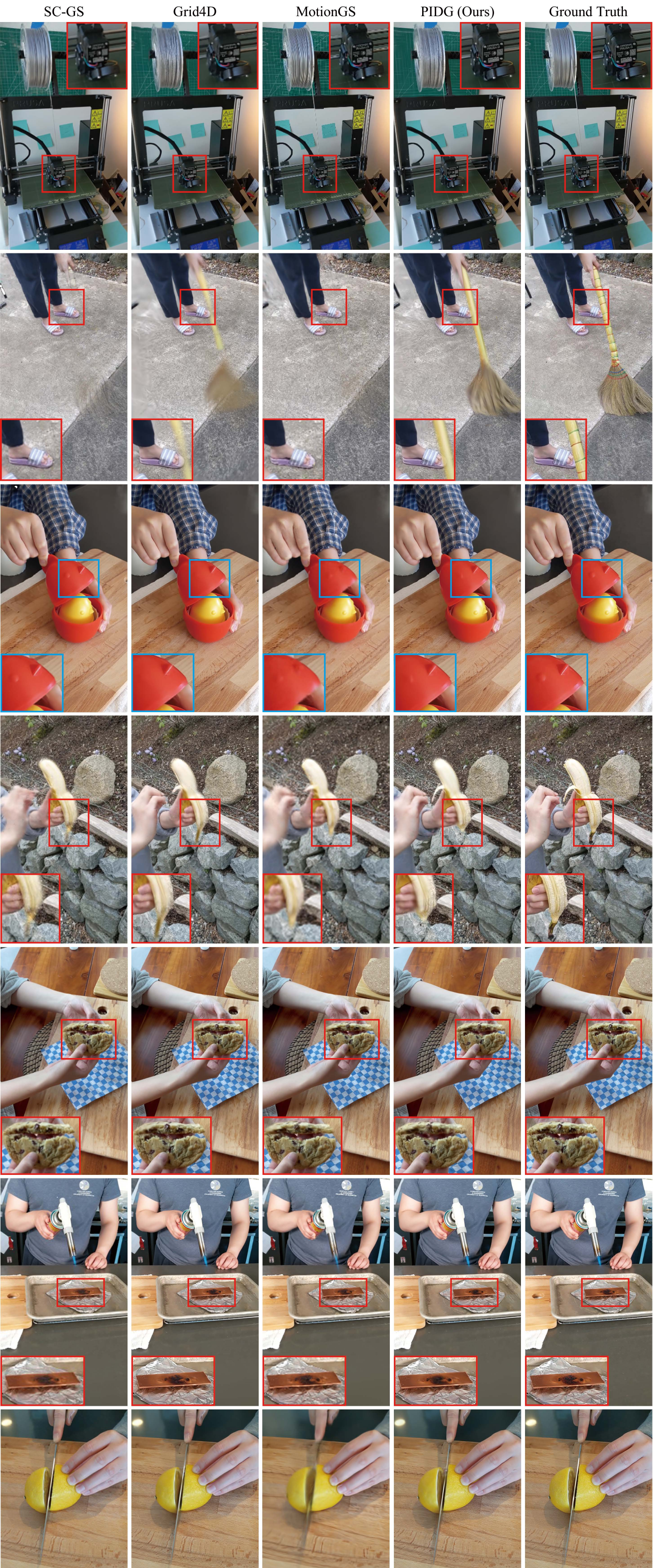} % Reduce the figure size so that it is slightly narrower than the column. Don't use precise values for figure width.This setup will avoid overfull boxes.
\caption{\textbf{The comparison of dynamic novel-view synthesis visual results on the HyperNeRF real-world dataset}. Top to bottom: SC-GS~\cite{huang2024sc}, Grid4D~\cite{xugrid4d}, MotionGS~\cite{zhumotiongs}, PIDG (Ours), and Ground Truth.
Left to right: monocular videos from the 3D-Printer, Broom, Chicken, and Peel-Banana scenes in the vrig dataset, followed by large-deformation dynamic scenes Split-Cookie, Torchocolate, and Cut-Lemon.}
\label{hyper_all}
\end{figure}

The complete PIDG with $\mathcal{L}_{\rm LPFM}$ achieves the best or near-best performance in several cases, most visibly in Gaussian-flow visualizations, by producing more temporally coherent reconstructions. Nevertheless, in scenes with weak dynamics or inaccurate pretrained optical flow, the margin over baselines narrows, yet still brings positive effects in rendering.  Overall, our method effectively captures intrinsic physical properties and enhances the physical plausibility of real-world dynamic reconstruction. Although improvements in 2D quantitative metrics are less pronounced, particularly when compared to synthetic scenes with larger motion coverage and more uniform motion patterns, our analysis of pretrained optical flow, motion masks, velocity fields, and Gaussian flow alignment provides strong evidence that PIDG substantially improves 3D motion coherence, underscoring its effectiveness in enforcing physically consistent dynamics under complex real-world conditions.

\subsection{Detailed results on D-NeRF datasets.}

\begin{table*}[ht]
\centering
\resizebox{\textwidth}{!}{%
\begin{tabular}{@{}l*{12}{c}@{}}
\toprule
\multirow{2}{*}{Method}
  & \multicolumn{3}{c}{Hell Warrior}
  & \multicolumn{3}{c}{Mutant}
  & \multicolumn{3}{c}{Hook}
  & \multicolumn{3}{c}{Bouncing Balls}\\
\cmidrule(lr){2-4}\cmidrule(lr){5-7}\cmidrule(lr){8-10}\cmidrule(lr){11-13}
  & PSNR $\uparrow$ & SSIM $\uparrow$ & LPIPS $\downarrow$
  & PSNR $\uparrow$ & SSIM $\uparrow$ & LPIPS $\downarrow$
  & PSNR $\uparrow$ & SSIM $\uparrow$ & LPIPS $\downarrow$
  & PSNR $\uparrow$ & SSIM $\uparrow$ & LPIPS $\downarrow$\\
\midrule
D-NeRF~\cite{pumarola2021d}           & 24.06 & 0.944 & 0.071 & 30.31 & 0.967 & 0.039 & 29.02 & 0.960 & 0.055 & 38.17 & 0.989 & 0.032\\
TiNeuVox~\cite{fang2022fast}         & 27.10 & 0.964 & 0.077 & 31.87 & 0.961 & 0.047 & 30.61 & 0.960 & 0.059 & 40.23 & 0.993 & 0.042\\
Tensor4D~\cite{shao2023tensor4d}         & 31.26 & 0.925 & 0.074 & 29.11 & 0.945 & 0.060 & 28.63 & 0.943 & 0.064 & 24.47 & 0.962 & 0.044\\
K-Planes~\cite{fridovich2023k}         & 24.58 & 0.952 & 0.082 & 32.50 & 0.971 & 0.036 & 28.12 & 0.949 & 0.066 & 40.05 & 0.993 & 0.032\\
3D-GS~\cite{kerbl20233d}            & 29.89 & 0.916 & 0.106 & 24.53 & 0.934 & 0.058 & 21.71 & 0.888 & 0.103 & 23.20 & 0.959 & 0.060\\
4D-GS ~\cite{wu20244d}           & 38.52 & 0.975 & 0.052 & 38.80 & 0.986 & 0.021 & 33.83 & 0.973 & 0.034 & 37.69 & 0.992 & 0.015\\
D-3DGS ~\cite{yang2024deformable}          & 41.28 & 0.987 & 0.025 & 42.11 & 0.994 & 0.007 & 36.76 & 0.985 & 0.017 & 41.54 & 0.995 & 0.009\\
GaussianPredict ~\cite{zhao2024gaussianprediction} & 41.73 & 0.987 & 0.021 & 42.90 & 0.995 & \underline{0.005} & 37.44 & 0.987 & 0.014 & 41.57 & 0.995 & 0.009\\
SC-GS  ~\cite{huang2024sc}          & 42.19 & 0.989 & \underline{0.019} & 43.43 & \underline{0.996} & \textbf{0.004} & 38.19 & 0.990 & \underline{0.011} & 41.59 & 0.995 & 0.009\\
Grid4D ~\cite{xugrid4d}          & 42.85 & 0.991 & \textbf{0.015} & 43.94 & \underline{0.996} & \textbf{0.004} & 38.89 & 0.990 & \textbf{0.009} & 42.62 & \underline{0.996} & \underline{0.008}\\
MoDec-GS ~\cite{kwak2025modec} & 24.41 & 0.954 & 0.061 & 31.01 & 0.968 & 0.036 & 26.90 & 0.944 & 0.058 & 37.68 & 0.991 & 0.011\\
D-2DGS ~\cite{zhang2024dynamic2dgaussiansgeometrically} & 41.65 & \textbf{0.993} & 0.021 & 42.48 & \textbf{0.999} & 0.006 & 38.53 & \textbf{0.996} & 0.012 & 33.75 & 0.984 & 0.036\\
\specialrule{0.1em}{0.5pt}{0.5pt}
\addlinespace[2pt]
GaussianPredict  $+ \ \mathcal{L}_{\rm CMR}$ & 42.02 & 0.988 & 0.020 & 43.22 & \underline{0.996} & \underline{0.005} & 37.83 & 0.988 & 0.012 & 41.94 & \underline{0.996} & \underline{0.008} \\
SC-GS $+ \ \mathcal{L}_{\rm CMR}$ & 42.25 & 0.990 & \underline{0.019} & 43.45 & \underline{0.996} & \textbf{0.004} & 39.08 & 0.990 & \underline{0.011} & 42.22 & \textbf{0.997} & \underline{0.008}\\
Grid4D $+ \ \mathcal{L}_{\rm CMR}$ & \underline{43.03} & \underline{0.992} & \textbf{0.015} & \underline{44.07} & \underline{0.996} & \textbf{0.004} & \underline{39.12} & \underline{0.991} & \textbf{0.009} & \underline{42.63} & \underline{0.996} & \textbf{0.007}\\
PIDG ($w/o\ \mathcal{L}_{\rm LPFM}$) & \textbf{43.08} & \underline{0.992} & \textbf{0.015} & \textbf{44.08} & \underline{0.996} & \textbf{0.004} & \textbf{39.15} & \underline{0.991} & \textbf{0.009} & \textbf{42.69} & \underline{0.996} & \textbf{0.007}\\
\midrule
\multirow{2}{*}{Method}
  & \multicolumn{3}{c}{T-Rex}
  & \multicolumn{3}{c}{Stand Up}
  & \multicolumn{3}{c}{Jumping Jacks}
  & \multicolumn{3}{c}{Average}\\
\cmidrule(lr){2-4}\cmidrule(lr){5-7}\cmidrule(lr){8-10}\cmidrule(lr){11-13}
  & PSNR $\uparrow$ & SSIM $\uparrow$ & LPIPS $\downarrow$
  & PSNR $\uparrow$ & SSIM $\uparrow$ & LPIPS $\downarrow$
  & PSNR $\uparrow$ & SSIM $\uparrow$ & LPIPS $\downarrow$
  & PSNR $\uparrow$ & SSIM $\uparrow$ & LPIPS $\downarrow$\\
\midrule
D-NeRF ~\cite{pumarola2021d}          & 30.61 & 0.967 & 0.054 & 33.13 & 0.978 & 0.036 & 32.70 & 0.978 & 0.039 & 31.14 & 0.969 & 0.046\\
TiNeuVox  ~\cite{fang2022fast}       & 31.25 & 0.967 & 0.048 & 34.61 & 0.980 & 0.033 & 33.49 & 0.977 & 0.041 & 32.74 & 0.972 & 0.050\\
Tensor4D ~\cite{shao2023tensor4d}        & 23.86 & 0.935 & 0.054 & 30.56 & 0.958 & 0.036 & 24.20 & 0.925 & 0.067 & 27.44 & 0.942 & 0.057\\
K-Planes  ~\cite{fridovich2023k}       & 30.43 & 0.974 & 0.034 & 33.10 & 0.979 & 0.031 & 31.11 & 0.971 & 0.047 & 31.41 & 0.970 & 0.047\\
3D-GS   ~\cite{kerbl20233d}         & 21.93 & 0.954 & 0.049 & 21.91 & 0.930 & 0.079 & 20.64 & 0.930 & 0.083 & 23.40 & 0.930 & 0.077\\
4D-GS     ~\cite{wu20244d}       & 33.60 & 0.986 & 0.019 & 40.43 & 0.989 & 0.016 & 35.59 & 0.984 & 0.021 & 36.92 & 0.984 & 0.026\\
D-3DGS   ~\cite{yang2024deformable}        & 37.73 & 0.993 & 0.010 & 43.73 & 0.994 & 0.009 & 37.51 & 0.990 & 0.014 & 40.09 & 0.991 & 0.013\\
GaussianPredict ~\cite{zhao2024gaussianprediction} & 37.39 & 0.993 & 0.011 & 45.09 & 0.995 & 0.006 & 37.93 & 0.991 & 0.010 & 40.58 & 0.992 & 0.011\\
SC-GS ~\cite{huang2024sc}           & 39.53 & \underline{0.994} & \underline{0.009} & \underline{46.72} & 0.997 & \underline{0.004} & 39.34 & 0.992 & \underline{0.008} & 41.65 & 0.993 & \textbf{0.008}\\ 
Grid4D   ~\cite{xugrid4d}        & \textbf{40.01} & \textbf{0.995} & \textbf{0.008} & 46.28 & 0.997 & \underline{0.004} & 39.37 & \underline{0.993}  & \underline{0.008} & 42.00 & \underline{0.994} & \textbf{0.008}\\
MoDec-GS ~\cite{kwak2025modec} & 30.29 & 0.972 & 0.038 & 32.46 & 0.978 & 0.029 & 31.68 & 0.976 & 0.033 & 30.63 & 0.969 & 0.038\\
D-2DGS ~\cite{zhang2024dynamic2dgaussiansgeometrically} & 21.12 & 0.923 & 0.056 & 45.76 & \textbf{0.999} & 0.005 & 38.99 & \textbf{0.997} & 0.011 & 37.47 & 0.984 & 0.021 \\

\specialrule{0.1em}{0.5pt}{0.5pt}
\addlinespace[2pt]
GaussianPredict  $+ \ \mathcal{L}_{\rm CMR}$ & 37.78 & 0.993 & 0.011 & 45.45 & 0.996 & 0.005 & 38.26 & 0.991 & 0.009 & 40.93 & 0.993 & \underline{0.010}\\
SC-GS $+ \ \mathcal{L}_{\rm CMR}$ & 39.47 & 0.993 & 0.010 & \textbf{46.91} & \textbf{0.999} & \underline{0.004} & \underline{39.51} & \underline{0.993}  & \underline{0.008} & 41.85 & \underline{0.994} & \textbf{0.008}\\
Grid4D $+ \ \mathcal{L}_{\rm CMR}$  & 39.85 & \textbf{0.995} & \textbf{0.008} & 46.55 & \underline{0.998} & \textbf{0.003} & 39.49 & \underline{0.993}  & \textbf{0.007} & \underline{42.10} & \textbf{0.995} & \textbf{0.008}\\
PIDG ($w/o\  \mathcal{L}_{\rm LPFM}$)  & \underline{39.88} & \textbf{0.995} & \textbf{0.008} & 46.57 & \underline{0.998} & \textbf{0.003} & \textbf{39.53} & \underline{0.993} & \textbf{0.007} & \textbf{42.14} & \textbf{0.995} & \textbf{0.008}\\
\bottomrule
\end{tabular}}
\caption{\textbf{Quantitative dynamic novel view synthesis results on D-NeRF dataset.} We also tested the improvement in dynamic novel view synthesis by introducing a time-evolving material field constrained solely by \(\mathcal{L}_{\rm CMR}\) into GaussianPredict, Grid4D, and PIDG. We report PSNR, SSIM, and LPIPS‑VGG. The best results are in \textbf{bold} and the second‑best are \underline{underlined}.}
\label{exp:dnerf}
\end{table*}

\begin{table*}[!b]
\centering
\resizebox{\textwidth}{!}{%
  % \footnotesize
  \begin{tabular}{lcccccccc}
    \toprule
    & \multicolumn{2}{c}{T-Rex} 
    & \multicolumn{2}{c}{Jumping Jacks} 
    & \multicolumn{2}{c}{Bouncing Balls} 
    & \multicolumn{2}{c}{Hell Warrior} \\
    \cmidrule(lr){2-3} \cmidrule(lr){4-5} \cmidrule(lr){6-7} \cmidrule(lr){8-9}
    \multirow{-2}{*}{Method} 
      & PSNR & SSIM 
      & PSNR & SSIM 
      & PSNR & SSIM 
      & PSNR & SSIM \\
    \midrule

    4D-GS 
      & 20.72 & 0.940 
      & 20.28 & \underline{0.918}  
      & 29.42 & 0.975 
      & 31.48 & 0.927 \\

    D-3DGS 
      & 20.81 & 0.943 
      & 20.21 & 0.915 
      & 28.90 & 0.978 
      & 29.82 & 0.914 \\
       
    GaussianPredict (MLP) 
      & 21.51 & 0.944 
      & 20.68 & \textbf{0.919} 
      & 29.58 & \underline{0.982} 
      & 29.99 & 0.918 \\

    GaussianPredict (GCN) 
      & 21.09 & 0.941 
      & 20.51 & \underline{0.918} 
      & 26.63 & 0.971 
      & 30.75 & 0.928 \\

    GaussianPredict (MLP) + $\mathcal{L}_{\rm CMR}$  
      & 21.62 ($\uparrow$0.51\%) & 0.946 ($\uparrow$0.21\%) 
      & \underline{20.70} ($\uparrow$0.10\%) & \textbf{0.919} ($\uparrow$0.00\%) 
      & \underline{29.69} ($\uparrow$0.37\%) & \textbf{0.986} ($\uparrow$0.41\%) 
      & 31.02 ($\uparrow$3.43\%) & 0.932 ($\uparrow$1.53\%) \\

    GaussianPredict (GCN) + $\mathcal{L}_{\rm CMR}$  
      & 21.40 ($\uparrow$1.47\%) & 0.945 ($\uparrow$0.43\%) 
      & 20.64 ($\uparrow$0.63\%) & 0.916 ($\downarrow$0.22\%) 
      & 27.52 ($\uparrow$3.34\%) & 0.974 ($\uparrow$0.31\%) 
      & 31.46 ($\uparrow$2.31\%) & 0.932 ($\uparrow$0.63\%) \\
    \specialrule{0.1em}{0.5pt}{0.5pt}
    \addlinespace[2pt]
    PIDG ($w/o\ (\mathcal{L}_{\rm CMR}+\mathcal{L}_{\rm LPFM})$) 
      & \underline{22.02} & \underline{0.947} 
      & 20.53 & \underline{0.918} 
      & 29.34 & 0.978 
      & \underline{31.83} & \underline{0.937} \\

    PIDG ($w/o\  \mathcal{L}_{\rm LPFM}$)
      & \textbf{22.11} & \textbf{0.948} 
      & \textbf{20.71} & \textbf{0.919} 
      & \textbf{29.78} & 0.980 
      & \textbf{32.37} & \textbf{0.939} \\

    \midrule[\heavyrulewidth]

    & \multicolumn{2}{c}{Mutant} 
    & \multicolumn{2}{c}{Stand Up} 
    & \multicolumn{2}{c}{Hook} 
    & \multicolumn{2}{c}{Average} \\ 
    \cmidrule(lr){2-3} \cmidrule(lr){4-5} \cmidrule(lr){6-7} \cmidrule(lr){8-9}
    \multirow{-2}{*}{Method} 
      & PSNR & SSIM 
      & PSNR & SSIM 
      & PSNR & SSIM 
      & PSNR & SSIM \\
    \midrule

    4D-GS 
      & 24.61 & 0.927 
      & 22.25 & 0.914 
      & 23.92 & 0.904 
      & 24.67 & 0.929 \\

    D-3DGS 
      & 24.32 & 0.930 
      & 21.38 & 0.913 
      & 21.41 & 0.887 
      & 23.84 & 0.926 \\

    GaussianPredict (MLP) 
      & 25.05 & 0.936 
      & 23.04 & 0.925 
      & 22.60 & 0.897 
      & 24.64 & 0.932 \\

    GaussianPredict (GCN) 
      & \underline{28.16} & \underline{0.956} 
      & \underline{25.96} & \underline{0.940} 
      & 23.42 & 0.909 
      & 25.22 & \underline{0.938} \\

    GaussianPredict (MLP) + $\mathcal{L}_{\rm CMR}$  
      & 25.79 ($\uparrow$2.95\%) & 0.939 ($\uparrow$0.32\%) 
      & 23.22 ($\uparrow$0.78\%) & 0.926 ($\uparrow$0.11\%) 
      & 22.75 ($\uparrow$0.66\%) & 0.903 ($\uparrow$0.67\%) 
      & 24.97 ($\uparrow$1.36\%) & 0.936 ($\uparrow$0.46\%) \\

    GaussianPredict (GCN) + $\mathcal{L}_{\rm CMR}$  
      & \textbf{29.72} ($\uparrow$5.54\%) & \textbf{0.962} ($\uparrow$1.06\%) 
      & \textbf{27.62} ($\uparrow$6.39\%) & \textbf{0.950} ($\uparrow$0.55\%) 
      & \textbf{23.94} ($\uparrow$2.22\%) & \textbf{0.914} ($\uparrow$0.46\%) 
      & \textbf{26.04} ($\uparrow$3.27\%) & \textbf{0.942} ($\uparrow$0.46\%) \\
        \specialrule{0.1em}{0.5pt}{0.5pt}
    \addlinespace[2pt]

    PIDG ($w/o\ (\mathcal{L}_{\rm CMR}+\mathcal{L}_{\rm LPFM})$) 
      & 25.61 & 0.933 
      & 23.16 & 0.925 
      & 23.65 & 0.912 
      & 25.16 & 0.936 \\

    PIDG ($w/o\  \mathcal{L}_{\rm LPFM}$) 
      & 25.85 & 0.943 
      & 23.45 & 0.927 
      & \underline{23.93} & \underline{0.913} 
      & \underline{25.46} & \underline{0.938} \\
    \bottomrule
  \end{tabular}%
}
\caption{\textbf{Quantitative motion prediction results comparison on D-NeRF dataset}. We also tested the improvement in future prediction by introducing a time-evolving material field constrained solely by \(\mathcal{L}_{\rm CMR}\) into GaussianPredict, evaluating both MLP and GCN modes. The best results are in \textbf{bold} and the second‑best are \underline{underlined}. Notably, encoding continuous velocities through the time-evolving material field exhibits consistently positive effects across different baselines.}
\label{tab:D-NeRF_extrapolate}
\end{table*}

We further evaluate our method on the D-NeRF synthetic dataset~\cite{pumarola2021d}. Because the camera viewpoints are not temporally continuous, reliable optical flow estimation is challenging. Consequently, we introduce a time-evolving material field constrained solely by $\mathcal{L}_{\rm CMR}$ and plug it into several state-of-the-art pipelines as a plug-and-play module to assess its impact on dynamic novel view synthesis and future prediction.

As reported in Tab.~\ref{exp:dnerf}, quantitative results for dynamic novel view synthesis show that injecting the $\mathcal{L}_{\rm CMR}$-only material field into GaussianPredict~\cite{zhao2024gaussianprediction}, SC-GS~\cite{huang2024sc}, and Grid4D~\cite{xugrid4d} yields consistent improvements across most synthetic scenes. The gains are smaller on T-Rex and Stand Up, where the underlying continuum mechanics are more intricate. Note that scenes in D-NeRF mostly contain only the moving object of interest, so our static-dynamic decoupling mainly regularizes the rendering loss via a mask-based region-of-interest constraint, and optical flow cannot be leveraged as a proxy supervision to further constrain the time-evolving material field due to the discontinuous camera viewpoints. We also observe that GaussianPredict and SC-GS heavily rely on randomness in initialization and training, resulting in high variance across multiple runs, which can obscure consistent performance trends. Nevertheless, we observe that our proposed module consistently brings relative improvements across these runs, indicating its robustness and effectiveness. Ultimately, PIDG ($w/o~\mathcal{L}_{\rm LPFM}$) achieves the best average score, indicating that our canonical hash-space decoupling and time-evolving material field remain effective even without pretrained optical flow as a data-fitting term.

\begin{figure}[t]
\centering
\includegraphics[width=1.0\columnwidth]{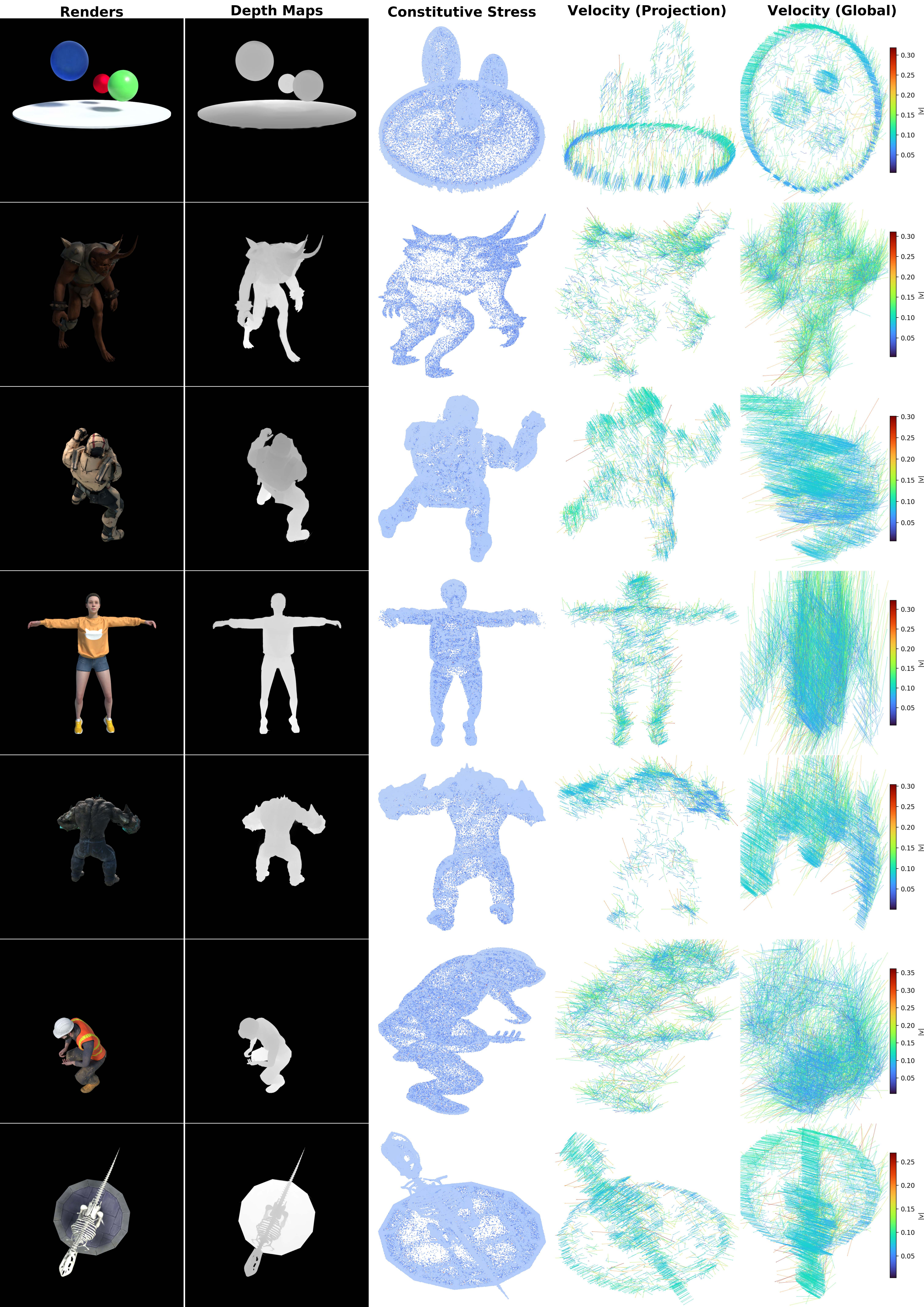} % Reduce the figure size so that it is slightly narrower than the column. Don't use precise values for figure width.This setup will avoid overfull boxes.
\caption{\textbf{Dynamic reconstruction and modeling results of our PIDG method in D-NeRF datasets.} From left to right, we show the RGB rendering images at a specified view and time, the predicted depth maps, the magnitude of the $x$-component of displacement associated with the constitutive stress, the projected velocity quiver of the deformation field and the velocity quiver in canonical space.}
\label{exp_dnerf_nvs}
\end{figure}

As illustrated in Fig.~\ref{exp_dnerf_nvs}, PIDG simultaneously models spatiotemporal structure and intrinsic physical properties, providing not only satisfactory RGB and depth renderings but also predicting velocity and stress fields through the material field. The velocity field reveals coherent motion among dynamic particles, suggesting that most particles evolve under the Cauchy momentum framework, with the dynamic particles exhibiting consistent velocity directions and magnitudes. For quantitative future prediction results on D-NeRF in Tab.~\ref{tab:D-NeRF_extrapolate}, GaussianPredict distils motion into keypoints and models future dynamics using either MLP extrapolation or GCN-based training. Following its original protocol, we train on frames from $[0, 0.8]$ and evaluate on $[0.8, 1.0]$. Incorporating the $\mathcal{L}_{\rm CMR}$-only material field strengthens the consistency of Gaussian particles as they deform, improving both the MLP and GCN variants, this addition mitigates mode collapse and produces motion amplitudes closer to the ground truth, demonstrating the generalization ability of our framework for physical modeling in dynamic scenes.  Moreover, when performing extrapolation with PIDG ($w/o~(\mathcal{L}_{\rm CMR}+\mathcal{L}_{\rm LPFM})$) and PIDG ($w/o~\mathcal{L}_{\rm LPFM}$), we also observe clear gains brought by the time-evolving material field. Nevertheless, because the GCN branch in GaussianPredict preserves richer dynamic cues from the distilled keypoints during future prediction, GaussianPredict (GCN) + $\mathcal{L}_{\rm CMR}$ attains the best average performance. It is worth noting that although direct extrapolation using the 3DGS representation, either through MLP-based or GCN-based methods, performs well in relatively simple synthetic dynamic scenes, it still suffers from mode collapse when facing more complex motions or real-world scenarios. In this work, we utilize the D-NeRF dataset to demonstrate the positive effect of introducing the time-evolving material field for future prediction, highlighting its potential applicability to broader dynamic modeling tasks.

Based on extensive experiments across the PIDG physics-driven synthetic dataset, D-NeRF synthetic dataset, and HyperNeRF real-world dataset, our approach consistently achieves high-fidelity dynamic novel view synthesis through physics-informed Gaussian representation, empirically validating superior cross-scene generalization to various physical phenomena, enhanced spatiotemporal coherence via aligned Gaussian flows, and generalizable modeling of physical properties like stress and velocity fields, while robustly improving reconstruction quality under diverse challenging conditions, thereby systematically strengthening the methodological claims in the main manuscript.

\section{Additional Related Work}

\subsection{Dynamic Reconstruction Methods}

\subsubsection{Learning-based Dynamic Reconstruction.}

Early methods relied on explicit geometry~\cite{broxton2020immersive, newcombe2015dynamicfusion, orts2016holoportation}. Recent NeRF-based approaches map dynamic scenes to a canonical space~\cite{mildenhall2020nerf, du2021neural, park2021hypernerf, park2021nerfies, wang2023tracking}, or leverage time-varying NeRF~\cite{li2021neural, xian2021space, gao2021dynamic, cao2023hexplane} and explicit representations~\cite{fridovich2023k, wang2023mixed}, but they incur high computational cost from repeated sampling and MLP queries. To improve efficiency, 3DGS~\cite{kerbl20233d} has been adopted for 4D scene reconstruction. DynamicGS~\cite{luiten2024dynamic} employs incremental modeling, while deformation-field methods combine Gaussians with implicit deformation fields~\cite{wu20244d, yang2024deformable, huang2024sc, bae2024per}. SC-GS~\cite{huang2024sc} uses sparse controls to accelerate dynamic reconstruction, but its random initialization and low-rank assumption will lose motion patterns. Grid4D~\cite{xugrid4d} embeds motion in a hash-coded space to better preserve patterns, yet it still struggles to learn dynamics uniformly.

% Early methods used explicit geometry~\cite{broxton2020immersive,newcombe2015dynamicfusion,orts2016holoportation}. NeRF-based approaches either map dynamic scenes into a canonical space~\cite{mildenhall2020nerf,park2021nerfies} or represent time-varying geometry with implicit or explicit fields~\cite{li2021neural,fridovich2023k,park2021hypernerf,pumarola2021d}, but both suffer high sampling and MLP costs. For greater efficiency, 3DGS~\cite{kerbl20233d} enables fast 4D reconstruction, DynamicGS~\cite{luiten2024dynamic} adds Gaussians incrementally, and deformation-field methods warp Gaussians via learned fields~\cite{wu20244d,yang2024deformable,zhao2024gaussianprediction}. SC-GS~\cite{huang2024sc} accelerates reconstruction with sparse controls at the expense of motion detail, while Grid4D~\cite{xugrid4d} uses hash coding to better preserve dynamics, yet it still struggles to learn dynamics uniformly.

\subsubsection{Optical Flow Supervision.}

Because RGB-only supervision will miss potential dynamics, recent work has turned to optical-flow supervision. GaussianFlow~\cite{gao2024gaussianflow} incorporates supervision from optical flow fields; MotionGS~\cite{zhumotiongs} disentangles object motion from camera motion; MAGS~\cite{guo2024motion} encodes Gaussian particles and aligns their velocity embeddings with pixel-level optical flow. Despite impressive rendering quality in specific scenarios, these methods lack a deep understanding of the physical constitutive laws of motion. Consequently, they struggle to generalize across diverse dynamic scenes.

\begin{figure*}[ht]
\centering
\includegraphics[width=1.8\columnwidth]{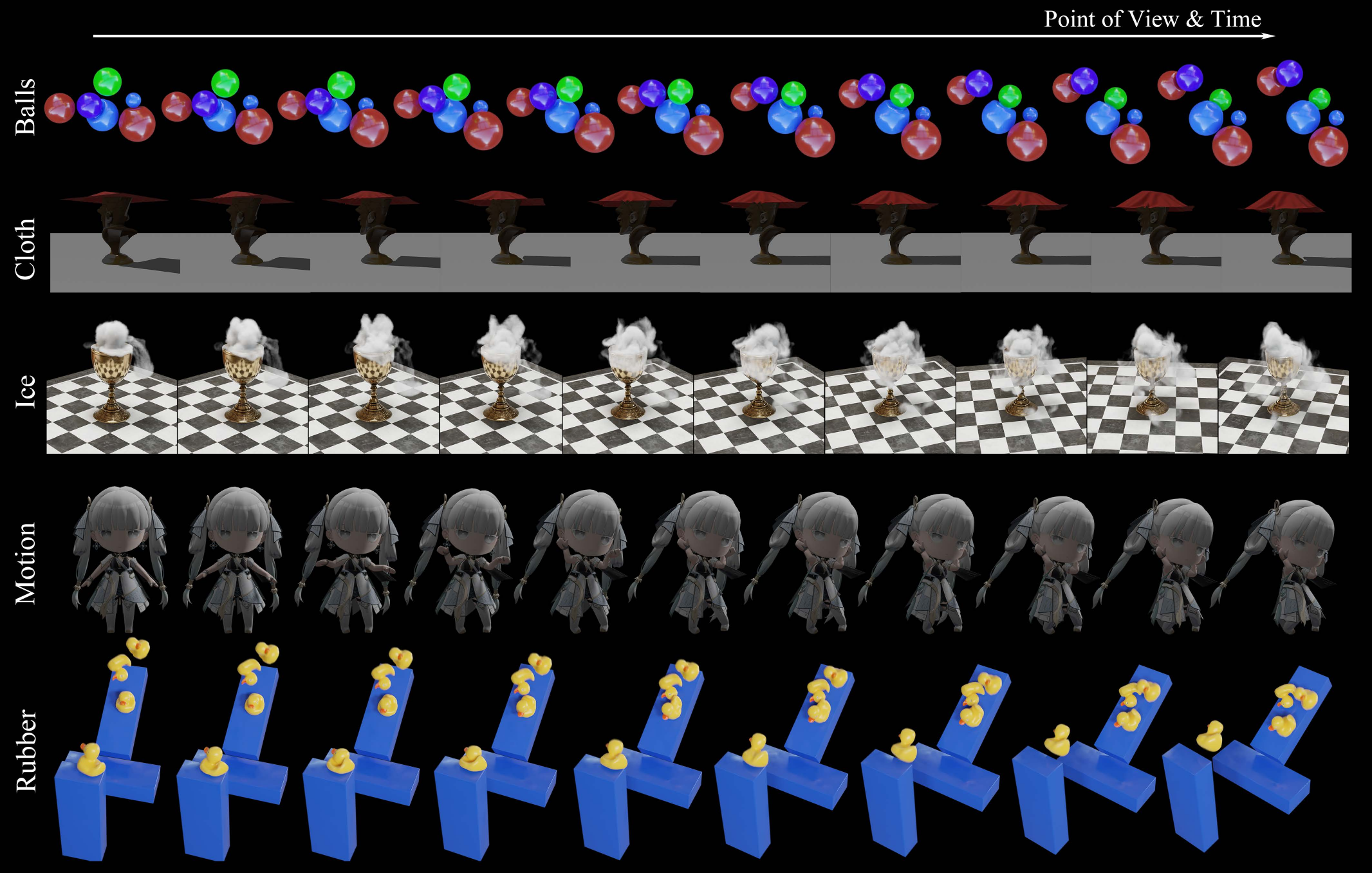} % Reduce the figure size so that it is slightly narrower than the column. Don't use precise values for figure width.This setup will avoid overfull boxes.
\caption{\textbf{Sample continuous video frames from our physics-driven PIDG synthetic dataset.} Different from D-NeRF's random-viewpoint synthetic data, PIDG features continuous camera trajectories that strictly adhere to monocular video constraints, enabling reliable optical flow extraction and thus significantly broadening its applicability.}
\label{pidg_datasets_continuous}
\end{figure*}

\subsection{Physics-Embedded Novel View Synthesis}
\subsubsection{Towards Physically Consistent Modeling.} 
Introducing physics into novel view synthesis offers a robust path for real-world model construction. Methods based on Material Point Methods (MPM)~\cite{zhang2025physdreamer,xie2024physgaussian,li2023pac,lin2025omniphysgs} leverage Eulerian grid stability and Lagrangian particle tracking to handle large deformations and intricate boundaries. Still, their reliance on grid discretization limits fluid-dynamic accuracy. FluidNexus~\cite{gao2025fluidnexus} combines novel-view video synthesis and physics-coupled particles for single-video 3D fluid reconstruction and prediction, but it relies on strict boundary conditions and pretrained view synthesis models. Spring-Gaus ~\cite{zhong2025reconstruction} and PhysTwin ~\cite{jiang2025phystwinphysicsinformedreconstructionsimulation} augment Gaussian kernels with spring-mass dynamics, enabling basic physical simulation but remaining tied to fixed material properties and boundary conditions. 

Methods based on Physics-Informed Neural Networks (PINNs)~\cite{chu2022physics,wang2024physics,yu2024inferring} embed the Navier-Stokes equations for smoke reconstruction, yet their fixed constitutive laws impede adaptation to varied physics. Moreover, these approaches simply incorporate partial differential equations as regularization terms in the loss function, without considering the constraints that PINNs should ensure the solution converges to physically meaningful boundary conditions or analytical solutions. Specifically, they lack proper data fitting terms that would guarantee physical consistency.  Despite these advances, a universal framework for monocular dynamic novel-view synthesis tasks remains lacking. 

To address this gap, we propose Physics-Informed Deformable Gaussian Splatting (PIDG), which unifies dynamic reconstruction and physics embedding within a single, fully differentiable framework. Unlike prior deformation-field approaches, PIDG models each Gaussian as a Lagrangian particle whose trajectory and internal state evolve under learned physical constitutive laws, preserving fine-grained motion patterns without relying on fixed material properties or low-rank assumptions. The integration of a time-evolving material field and Lagrangian particle flow matching enables PIDG to achieve physically consistent and temporally coherent reconstructions that generalize across diverse dynamic scenes, outperforming prior Gaussian-based methods in motion fidelity and physical plausibility.

\section{Additional Conclusion}
\subsubsection{Discussion and Advocacy.} Through our investigation of dynamic novel-view synthesis in both synthetic and real-world environments, we have found that relying exclusively on 2D standard metrics such as PSNR, SSIM, and LPIPS is inadequate for capturing the reconstruction qualities we value most, namely 3D structural accuracy and physical consistency, which demand evaluation in higher dimensions. For example, although GaussianPredict~\cite{zhao2024gaussianprediction} achieves a higher PSNR than our PIDG method on the Dry Ice fluid simulation scene, a qualitative examination of the renderings shows that its handling of sparse, smoke-like dynamics is poor, producing numerous sharp, anisotropic artifacts. By depending on static components, it can nonetheless score well under purely visual benchmarks, obscuring its failure to maintain physical plausibility. We therefore advocate that the novel-view synthesis community establish more comprehensive evaluation protocols that balance 2D rendering quality with 3D reconstruction fidelity and physical consistency, for instance by incorporating geometric, temporal and physics-based metrics.

\subsubsection{Limitations.} Our particle-based motion modeling approach performs well in dynamic synthetic scenes with relatively simple motion patterns, enforcing physical consistency of Gaussian particles and demonstrating promise for virtual reality, game design and embodied intelligence applications. However, the optimization-based reconstruction pipeline remains computationally intensive, requiring hours to days for a full training run on moderate-scale scenes and substantial GPU memory, which hinders scalability to large-scale or real-time scenarios. In addition, although the PINN-based constitutive model provides effective regularization, it does not fully capture complex material behaviors such as nonlinear elastoplasticity or viscoelasticity. When applied to real-world dynamic scenes with more intricate particle trajectories and interactions, the time-evolving material field can ensure plausible motion but cannot guarantee complete consistency of physical properties.

\subsubsection{Future Work.} Our method also demonstrates potential for mapping real-world events into virtual environments and constructing world models. To achieve real-time, high-fidelity dynamic reconstruction, we plan to develop lightweight feed-forward network architectures that avoid expensive optimization loops; integrate advanced physical simulation techniques such as hybrid finite-element/particle methods for richer material modeling; and combine scene segmentation with renormalization-group approaches to unify the motion of Gaussian particle ensembles, resulting in more coherent and physically faithful reconstructions across both synthetic and real-world dynamics.

\bibliography{aaai2026}

% \input{CheckList.tex}

% \appendix
% \section{Supplementary Material}
% \label{sec:sup_Appendix}
% \input{section/sup_Appendix}

\end{document}